\documentclass{article}
\usepackage{ijcai23}

\usepackage{times}
\usepackage{soul}
\usepackage{url}
\usepackage[hidelinks]{hyperref}
\usepackage[utf8]{inputenc}
\usepackage[small]{caption}
\usepackage{graphicx}
\usepackage{amsmath}
\usepackage{amsthm}
\usepackage{booktabs}
\usepackage{algorithm}
\usepackage{algorithmic}
\usepackage[switch]{lineno}

\usepackage{verbatim}
\usepackage{amsfonts,dsfont}
\usepackage{mathtools}
\usepackage[section]{placeins}
\usepackage{stfloats}

\allowdisplaybreaks

\newtheorem{theorem}{Theorem}
\newtheorem{fact}{Fact}

\newcommand{\alg}{\texttt{TP-UCB-FR-G}}
\newcommand{\ind}{\mathds{1}}   
\newcommand{\nArms}{K}
\theoremstyle{definition}
\newtheorem{definition}{Definition}

\DeclareMathOperator{\EX}{\mathbb{E}}
\DeclareMathOperator{\SIG}{\sum_{k=1}^{\alpha}
    k  B(k)}
\DeclareMathOperator{\SIGDOLLAR}{(\bar{R}^i)^2\sum_{k=1}^{\alpha}
    \left(B(k)\right)^2}
    
\DeclareMathOperator{\INDEXCOINCIDENCE}{\sum_{k=1}^{\alpha}
    \left(B(k)\right)^2}
    
\DeclareMathOperator{\lvalueleft}{\frac{2\phi \bar{R}^i \SIG}{\Delta_i}}
\DeclareMathOperator{\lvalueright}{+ \frac{4 \ln t\SIGDOLLAR }{\Delta^2_i} \left( 1+ \sqrt{1+ \frac{\Delta_i \phi \bar{R}^i \SIG}{\ln t \SIGDOLLAR }}\right)}

\DeclareMathOperator*{\argmax}{argmax}
\DeclareMathOperator*{\defined}{\coloneqq}

\newcommand{\regret}{\mathcal{R}}
\newcommand{\KLdiv}{\mathcal{KL}}
\usepackage{etoolbox}
\newtoggle{long-version}
\toggletrue{long-version}

\title{Multi-Armed Bandits with Generalized Temporally-Partitioned Rewards}
\author{
}
\author{
Ronald C. van den Broek$^1$
\and
Rik Litjens$^1$\and
Tobias Sagis$^1$\and
Luc Siecker$^1$\and
Nina Verbeeke$^1$\And
Pratik Gajane$^1$
\affiliations
$^1$Eindhoven University of Technology\\
\emails
r.c.v.d.broek@student.tue.nl,
r.litjens@student.tue.nl,
t.g.m.sagis@student.tue.nl,
l.r.siecker@student.tue.nl,
n.c.verbeeke@student.tue.nl,
p.gajane@tue.nl
}
\begin{document}
\maketitle
\begin{abstract}
Decision-making problems of sequential nature, where decisions made in the past may have an impact on the future, are used to model many practically important applications.
In some real-world applications, feedback about a decision is delayed and may arrive via partial rewards that are observed with different delays.
Motivated by such scenarios, we propose a novel problem formulation called multi-armed bandits with generalized temporally-partitioned rewards. To formalize how feedback about a decision is partitioned across several time steps, we introduce \textit{$\beta$-spread property}. We derive a lower bound on the performance of any uniformly efficient algorithm for the considered problem.
Moreover, we provide an algorithm called \alg \ and prove an upper bound on its performance measure. In some scenarios, our upper bound improves upon the state of the art. We provide experimental results 
validating the proposed algorithm and our theoretical results.
\end{abstract}


\section{Introduction}  
\label{sec:intro}
The classical multi-armed bandit (MAB, or simply bandit) problem is a framework to model sequential decision-making \cite{DBLP:journals/ftml/BubeckC12}. In a MAB problem, the learning agent is faced with a finite set of $\nArms$ arms, and a decision taken by the agent is symbolized by pulling an arm. Feedback about the decisions taken is available to the agent via numerical rewards. Multi-arm bandit literature typically focuses on scenarios where rewards are assumed to arrive immediately after pulling an arm. In contrast, the works on delayed-feedback bandits (e.g., \cite{Joulani2013,mandel2015queue}) assume a delay between pulling an arm and the observation of its corresponding reward. In those studies, the reward is assumed to be concentrated in a single round that is delayed. This setting can be extended by allowing the reward to be partitioned into partial rewards that are observed with different delays. This type of bandit problem, known as MAB with Temporally-Partitioned Rewards (TP-MAB), was introduced by \cite{Romano2022}.

In the TP-MAB setting, an agent will receive subsets of the reward over multiple rounds. The cumulative reward of an arm is the sum of the partial rewards obtained by pulling an arm. \cite{Romano2022} present $\alpha$-smoothness to characterize the reward structure. The $\alpha$-smoothness property states that the maximum reward in a group of consecutive partial rewards cannot exceed a fraction of the maximum reward (precise definition given in Definition~\ref{def:AlphaSmooth}).
However, the assumption of $\alpha$-smoothness does not fit well if the cumulative reward is not uniformly spread.
In this article, we introduce a more generalized way of formulating how an arm's delayed cumulative reward is spread across several rounds. 

\indent As a motivating application, consider websites (e.g., Coursera, Khan Academy, edX) that provide Massive Open Online Courses (MOOCs). Such websites aim to provide users with useful recommendations for courses. This problem can be modeled as a TP-MAP problem. A course, which consists of a series of video lectures, might be thought of as an arm. A course can be recommended to a user by an agent, which corresponds to pulling an arm. When the student follows a course, the agent can observe partial rewards (e.g., by checking the watch time retention).
In this setting, $\alpha$-smoothness rarely captures the actual cumulative reward distribution. Many students watch the video lectures at the beginning of a course but never finish the last few lectures, making the spread of partial rewards non-uniform. As a result, the existing work on delayed-feedback bandits and the algorithms proposed by \cite{Romano2022} may fail to recommend courses that are relevant for the user. Motivated by such scenarios, we investigate a more generalized way of formulating the reward structure.



\subsubsection*{Our Contributions}
\begin{enumerate}
    \item We introduce a novel MAB formulation with a generalized way of describing how an arm's delayed cumulative reward is distributed across rounds.
    \item We prove a lower bound on the performance measure of any uniformly efficient algorithm for the considered problem. 
    \item We devise an algorithm \alg \ and prove an upper bound on its performance measure. The proven upper bounds are tighter than the state of the art in some scenarios.
    \item We provide experimental results that validate the correctness of our theoretical results and the effectiveness of our proposed algorithm. 
\end{enumerate}

%

\section{Background and Related Work}
\begin{figure*}
  \centering
  \includegraphics[width=0.48\textwidth]{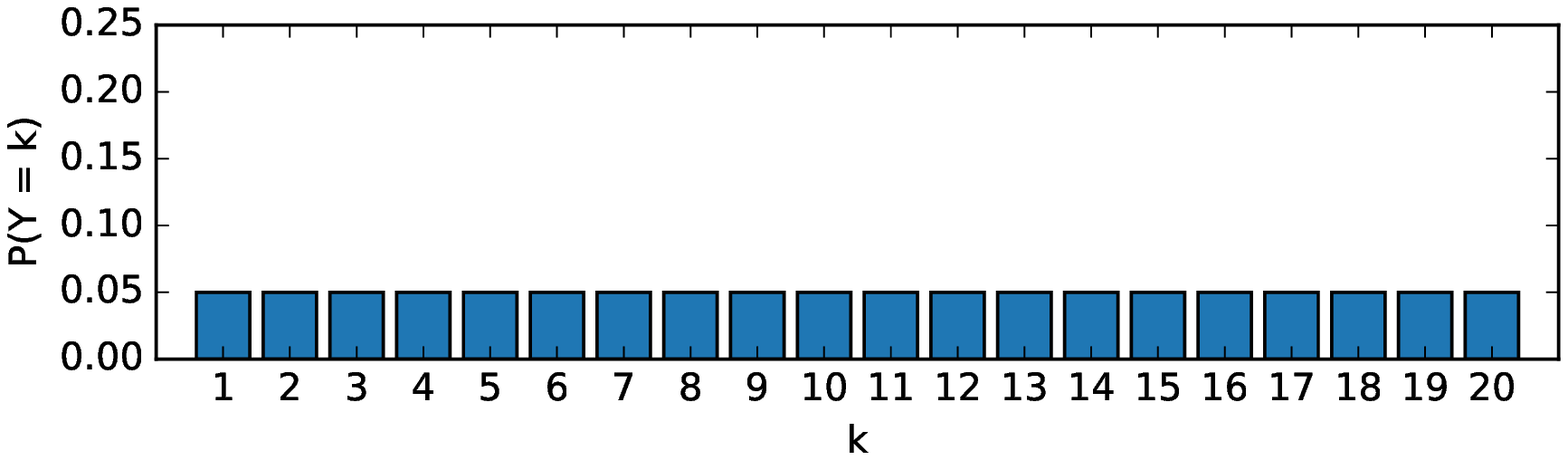}
  \hfill
  \includegraphics[width=0.48\textwidth]{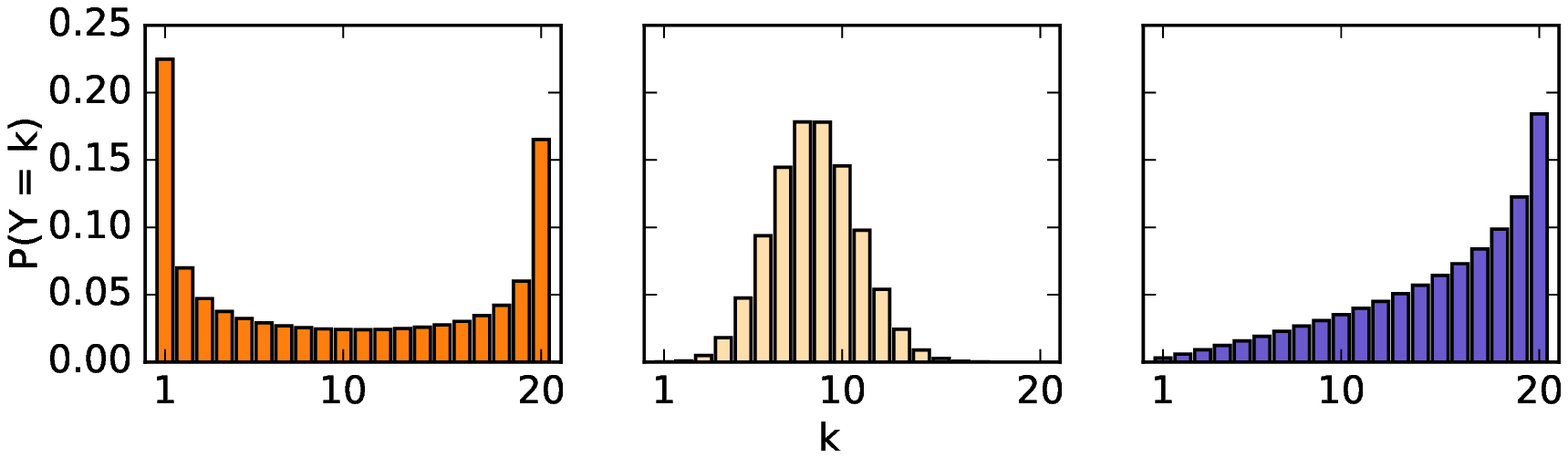}
  \caption{Probability distribution $\alpha$-smoothness (left) and possible probability distributions using our approach  $\beta$-spread (right)}
    \label{fig:alpha-beta-distr}
\end{figure*}

\label{Sec:Background}
Online learning with delayed feedback is a well-studied problem in the literature.
Owing to the space restrictions, a necessarily incomplete list of the works on this topic includes \cite{Weinberger2002,NIPS2011_f0e52b27,10.1007/11564089_31,MesterharmThesis,10.5555/2627435.2750368,NIPS2009_b55ec28c}. In the rest of this section, we focus on MAB with delayed feedback. 

\cite{Joulani2013} studied the \textit{non-anonymous} delayed feedback bandit problem and proposed a variant of the UCB algorithm \cite{Auer2002} as a solution. In \cite{Joulani2013}, it is assumed that knowledge of which action resulted in a specific delayed reward is available. 
\cite{Wang2021} extend this problem to contain \textit{anonymous} feedback and, in addition,  eliminate the need for accurate prior knowledge of the reward interval. Recently, a variety of delayed-feedback scenarios were studied in MAB settings different from ours, such as linear and contextual bandits \cite{ARYA2020108818,10.5555/3454287.3454754,pmlr-v119-vernade20a}, non-stationary bandits \cite{10.5555/3524938.3525839}. Furthermore, \cite{pike2018bandits} and \cite{JMLR:v23:21-1443} consider the case of delayed, aggregated, and anonymous feedback.



The majority of past research on the delayed MAB setting assumes that the entire reward of an arm is observed at once, either after some bounded delay \cite{Joulani2013,mandel2015queue} or after random delays from an unbounded distribution with finite expectation \cite{pmlr-v119-gael20a,vernade:hal-01545667}. Our article studies the setting in which the reward for an arm is spread over an interval with a finite maximum delay value. This is consistent with the applications that we aim to model, such as MOOC providers mentioned in Section \ref{sec:intro}.
To the best of our knowledge, \cite{Romano2022}, were the first to analyze this setting. 
They introduced the Multi-Armed Bandit with Temporally-Partitioned Rewards (TP-MAB) setting. In the TP-MAB setting, a stochastic reward that is received by pulling an arm is partitioned over partial rewards observed during a finite number of rounds followed by the pull. \cite{Romano2022} assume that the arm rewards follow $\alpha$-smoothness property (precise definition given in Definition~\ref{def:AlphaSmooth}).
%
%
As illustrated by the leftmost histogram in Figure \ref{fig:alpha-beta-distr}, if an arm's reward is $\alpha$-smooth, the probability of observing a partial reward in a group of consecutive rounds is uniformly partitioned.

While the study by \cite{Romano2022} provides promising results in the TP-MAB setting, it is based on the strong assumption that the $\alpha$-smoothness property holds. As a result, their proposed solutions are not suitable for a broader variety of scenarios where rewards are partitioned non-uniformly. 
As a remedy, we propose to use general distributions that can more accurately characterize how the received reward is partitioned. Consider a scenario in which additional information is available about how the cumulative reward is spread over the rounds. An example of such a scenario is a MOOC provider recommending courses to users, as described in Section \ref{sec:intro}. By generalizing the reward structure, our approach will be able to handle partitioned rewards in which the maximum reward per round is not partitioned uniformly across rounds, such as those shown on the right side of Figure \ref{fig:alpha-beta-distr}.


\cite{Romano2022} introduce two novel algorithms based on the UCB algorithm that leverage $\alpha$-smoothness property: \texttt{TP-UCB-EW} and \texttt{TP-UCB-FR}. Both algorithms take an assumed $\alpha$ value as input and use it to calculate confidence terms similar to UCB \cite{Auer2002}. \cite{Romano2022} showed that the \texttt{TP-UCB-EW} algorithm performs better with short time horizons $T$, whereas the \texttt{TP-UCB-FR} outperforms in the long run. Furthermore, they show that the cumulative regret of the \texttt{TP-UCB-FR} algorithm is greatly impacted by the assumed $\alpha$, whereas \texttt{TP-UCB-EW} only shows relatively minor changes in cumulative regret for different $\alpha$ values as input. Therefore, we believe that the setup of \texttt{TP-UCB-FR} is  most suitable for leveraging assumed distribution in a generalized setting too. Subsequently, we use \texttt{TP-UCB-FR} as a baseline for our proposed algorithm.

\section{Problem Formulation}
Consider a MAB problem with $K$ arms over a time horizon of $T$ rounds, where $K, T \in \mathbb{N}$. At every round $t \in \{1, 2, ..., T\}$ an arm from the set of arms $\{1, 2,...,K\}$ is pulled. 

The performance of an algorithm $\mathfrak{A}$ after $T$ time steps for the considered problem can be measured using expected \textit{regret} (or simply, regret) denoted as $\regret_T(\mathfrak{A})$.
\begin{definition}(Regret)
    The regret of an algorithm $\mathfrak{A}$ after $T$ time steps is 
    $\regret_T(\mathfrak{A}) \defined \mu^*T -  \sum_{i=1}^K \mu_i \cdot \mathbb{E}\left[N_i(T)\right]$,
where  $\mu^* \defined \max_{1 \leq i \leq K} \mu_i$ and  $N_i(T) =$ number of times an arm $i$ is selected till time $t$.
\end{definition}

The total reward is temporally partitioned over a set of rounds $T' = \{t, t+1,...,t+\tau_{\max}-1\}$. Let $x^i_{t, m} (m \in T')$ denote the partitioned reward that the learner receives at round $m$, after pulling the arm $i$ at round $t$. It is known to the agent which arm pull produced this reward. The cumulative reward is completely collected by the learner after a delay of at most $\tau_{\max}$. Each per-round reward $x^i_{t, m}$ is the realization of a random variable $X^i_{t, m}$ with support in $[0,  \overline{X}^i_{m}]$.
The cumulative reward collected by the learner from pulling arm $i$ at round $t$ is denoted by $r^i_t$ and it is the realization of a random variable $R^i_t$ such that $R^i_t := \sum_{n=t}^{t+ \tau_{\max} - 1} X^i_{t, n}$ with support $[0, \overline{R}^i]$. Straightforwardly, we observe that $\overline{R}^i := \sum_{n=t}^{\tau_{\max}} \overline{X}^i_{n}$.

\cite{Romano2022} have shown that, in practice, per-round rewards for an arm provide information on the cumulative reward of the arm. \cite{Romano2022} introduce an $\alpha$-smoothness property (defined in Definition \ref{def:AlphaSmooth}) that partitions the temporally-spaced rewards such that each partition corresponds to the sum of a set of consecutive per-round rewards. Formally, let $\alpha \in T'$ be such that $\alpha$ is a factor of $\tau_{\max}$. The cardinality of each partition, which we shall name `$z$-group' from now on, is denoted by $\phi := \frac{\tau_{\max}}{\alpha}$ with $\phi \in \mathbf{N}$. Similar to above, we can now define each $z$-group $z^i_{t, k}, k \in \{1, 2, ..., \alpha\}$ as the realization of a random variable $Z^i_{t, k}$ such that for every $k$:
\begin{equation}
    Z^{i}_{t, k} := \sum_{n=t+(k-1)\phi}^{t+k\phi-1} X^{i}_{t, n}
\end{equation}
such that $Z^i_{t, k}$ has support $[0, \overline{Z}^i_{\alpha, k}]$. 
\begin{definition}[$\alpha$-smoothness]
\label{def:AlphaSmooth}
For $\alpha \in \{1,...,\tau_{max}\}$, the reward is $\alpha$-smooth iff $\frac{\tau_{max}}{\alpha}$ 
$\in \mathbb{N}$ and for each $i \in \{1,...,K\}$ and $k \in \{1, 2, ..., \alpha\}$ the random variables $Z_{t,k}^i$ are independent and s.t. $\overline{Z}_{\alpha,k}^i = \overline{Z}_{\alpha}^i = \frac{\overline{R}^i}{\alpha}$.
\end{definition}


The $\alpha$-smoothness property ensures that all temporally-partitioned rewards contribute towards bounding the values of future rewards within the same window. 
If the $\alpha$-smoothness holds, then the maximum cumulative reward in a $z$-group $\overline{Z}_{\alpha, k}^i$ is equal for all $z$-groups $k \in \{1, 2, ..., \alpha\}$. Therefore, we can say that $\forall_k \in \{1, 2, ..., \alpha\}$, $\overline{Z}_{\alpha, k}^i = \overline{Z}_\alpha^i $. 

However, the assumption of $\alpha$-smoothness is unsuitable for scenarios in which the cumulative reward is not evenly partitioned across rounds. To fit such scenarios, the goal of this article is to generalize the spread of the rewards across $z$-groups. To that end, one has to eliminate the assumption that every $z$-group has an equal probability of attaining a partial reward. 
To accomplish this, we replace $\alpha$-smoothness with \textit{$\beta$-spread property} that allows for modeling scenarios in which the cumulative reward is not distributed uniformly across rounds i.e, a property that allows $\overline{Z}_{\alpha, k}^i$ to differ across $z$-groups.

\subsection{Our Solution approach: \texorpdfstring{$\beta$}{beta}-spread property}

Let us consider that for each partial reward, we have a discrete random variable $Y$ that represents the index of the $z$-group in which the partial reward will be observed. The set of all possible values that $Y$ can take is therefore $\{1,2,... \alpha \}$. Each possible value of $Y$ is associated with a certain probability that signifies the probability that the partial reward will be observed in $z$-group $k$. This means a non-uniform spread of rewards over the $z$-groups is possible. Formally, we define this concept of $\beta$-spread as follows:
\begin{definition}[$\beta$-spread]
\label{def:BetaSpread}Let $Y$ be a discrete random variable representing the index of the $z$-group in which a partial reward will be observed. $Y$ has a discrete probability distribution $\mathbb{F}$ with a finite domain of size $\alpha$ that is defined by a probability mass function denoted by $P_\mathbb{F}(k)$ for $k \in \{1,, \dots, \alpha\}$. We say that a reward of arm $i$ is $\beta$-spread if the partial rewards of arm $i$ are distributed according to an arbitrary discrete distribution $\mathbb{F}$. The upper bounds for the $z$-groups for every $k, i$ are defined as:
\begin{equation}
    \overline{Z}_{\alpha, k}^i=P_\mathbb{F}(k)\cdot\overline{R}^i
\end{equation}
\end{definition}

Based on prior information about how the cumulative reward is distributed over the rounds, the reward spread can be defined by any fitting probability distribution $\mathbb{F}$ as long as it adheres to the definition of $\beta$-spread. For example, beta-binomial distributions with a range starting from 1 and ending at $\alpha$ could be used to describe reward spread. The histograms on the right side of Figure \ref{fig:alpha-beta-distr} depict a few of these Beta-Binomial distributions with varying shape parameters, but the possibilities for choosing other parameters or distributions matching scenarios we aim to model are nearly inexhaustible.


\section{Lower Bound on Regret}

Using the $\beta$-spread property, we can derive the following lower bound for a  uniformly efficient policy i.e., any policy with regret in $o(T^x)$ with $x<1$.
\begin{theorem}
\label{Thm:LowerBound}
The regret of any uniformly efficient policy $\mathfrak{U}$ applied to the TP-MAB problem with the $\beta$-spread property after $T$ time steps is lower bounded as

\begin{align*}
    \lim \inf _{T \rightarrow+\infty} \frac{\regret_T(\mathfrak{U})}{\ln T} 
    &\geq \sum_{i: \mu_i<\mu^*}
    \frac{2}{(\alpha + 1)}\SIG \\
    & \qquad \cdot\alpha\INDEXCOINCIDENCE
    \frac{\Delta_i}
     {\alpha \KLdiv\left(\frac{\mu_i}{\bar{R}_{\max }}, \frac{\mu^*}{R_{\max }}\right)}
\end{align*}
where $\Delta_i \defined \mu^* - \mu_i$ and $\KLdiv(p, q) \defined $ Kullback-Leibler divergence between Bernoulli random variables with means p and q.
\end{theorem}

\subsubsection*{Comparison with the Lower Bound given by \protect\cite{Romano2022} }
By assuming $\alpha$-smoothness, \cite{Romano2022} proved the following lower bound for TP-MAB:

\begin{equation}
\label{eq:loweralphasmooth}
        \lim \inf _{T \rightarrow+\infty} \frac{\mathcal{R}_T(\mathfrak{U})}{\ln T} \geq \sum_{i: \mu_i<\mu^*} \frac{\Delta_i}{\alpha \KLdiv\left(\frac{\mu_i}{\bar{R}_{\max }}, \frac{\mu^*}{R_{\max }}\right)}.
\end{equation}

Note that our lower bound given in Theorem \ref{Thm:LowerBound} resolves to the lower bound given by \cite{Romano2022} in Eq.\eqref{eq:loweralphasmooth} in case of $\alpha$-smoothness. 
However, our lower bound for the considered problem setting is tighter when

\begin{equation*}
    \frac{2}{(\alpha + 1)}\SIG\cdot\alpha\INDEXCOINCIDENCE > 1.    
\end{equation*}

\subsubsection*{Proof Sketch for Theorem \ref{Thm:LowerBound}}
We start by constructing two MAB problem instances that call for different behaviors from the algorithm attempting to solve them.
Then, we use the change-of-distribution argument to show that any uniformly efficient algorithm cannot efficiently distinguish between these instances. 

\iftoggle{long-version}{The complete proof for Theorem \ref{Thm:LowerBound} is given in Appendix \ref{App:Proof:Thm:LowerBound}.}{Please consult the extended version of this article given in the supplementary material for the complete proof of Theorem \ref{Thm:LowerBound}.}

\section{Proposed Algorithm and Regret Upper Bound}
In this section, we propose an algorithm that makes use of the $\beta$-spread property in the TP-MAB setting and prove an upper bound on its regret.
\subsection{Proposed Algorithm: \alg}

Our proposed algorithm, called \alg, is an extension of the algorithm \texttt{TP-UCB-FR} given by \cite{Romano2022}.
 In \alg, the most significant modification is the confidence interval $c^i_{t-1}$ which is rigorously built to suit the $\beta$-spread property.
\begin{algorithm}[t]
\caption{TP-UCB-FR-G}\label{alg:fr}
\begin{algorithmic}[1]
    \STATE $\textbf{Input: } \alpha \in [\tau_{max}], \tau_{max} \in \mathbb{N}^*, B \sim Y$
    \FOR{$t \in \{1,...,K\}$}
        \STATE \text{Pull an arm $i_t=t$}
    \ENDFOR
    \FOR{$t \in \{K+1,...,T\}$}
        \FOR{$t \in \{1,...,K\}$}
            \STATE \text{Compute $\hat{R}_{t-1}^i$ and $c_{t-1}^i$} as in (\ref{eq:Rhat}) and (\ref{eq:c})
            \STATE \text{$u_{t-1}^i \leftarrow \hat{R}_{t-1}^i + c_{t-1}^i $}
        \ENDFOR
        \STATE \text{Pull arm $i_t= z = \argmax_{i \in [K]} u_{t-1}^i$ }
        \STATE \text{Observe $x_{h,t-h+1}^{i_h}$ for $h \in \{t-\tau_{max} +1, ..., t\}$}
    \ENDFOR
\end{algorithmic}
\end{algorithm}

As input, the algorithm takes a smoothness constant $\alpha \in [\tau_{max}]$, a maximum delay $\tau_{max}$ 
and a probability mass function $B$. The algorithm uses $B$ to be able to give a proper judgment of an arm before all the delayed partial rewards are observed. This is realized by replacing the not yet received partial rewards with fictitious realizations, or in other words, the expected estimated rewards.

At round $t$, the fictitious reward vectors are associated with each arm pulled in the span $H := \{t - \tau_{max} + 1, ..., t - 1 \}$. These fictitious rewards are denoted by $\tilde{\boldsymbol{x}}_h^i=\left[\tilde{x}_{h, 1}^i, \ldots, \tilde{x}_{h, \tau_{\max }}^i\right]$ with $h \in H$, where $\tilde{x}_{h, j}^i:=x_{h, j}^i$, if $h+j \leq t$ (the reward has already been seen), and $\tilde{x}_{h, j}^i=0$, if $h+j>t$ (the reward will be seen in the future). The corresponding fictitious cumulative reward is $\tilde{r}_h^i:=\sum_{j=1}^{\tau_{\max }} \tilde{x}_{h, j}^i$.

In the initialization phase of the algorithm (lines 2-4), each arm is pulled once. Later, at each time step $t$, the upper confidence bounds $u_{t-1}^i$ are determined for each arm $i$ by computing the estimated expected reward $\hat{R}_{t-1}^i$ and confidence interval $c_{t-1}^i$ using Eq. \eqref{eq:Rhat} and \eqref{eq:c} respectively. 
\begin{equation}
\label{eq:Rhat}
\hat{R}_{t-1}^i:=\frac{1}{N_i(t-1)}\left(\sum_{h=1}^{t-\tau_{\max }} r_h^i \mathbb{I}_{\left\{i_h=i\right\}}+\sum_{h \in H} \tilde{r}_h^i \mathbb{I}_{\left\{i_h=i\right\}}\right).
\end{equation}
where $N_i(t-1):=\sum_{h=1}^{t-1} \mathbb{I}_{\left\{i_h=i\right\}}$ is the number of times arm $i$ has been pulled up to round $t-1$.

\begin{align}
    c_{t-1}^i &= 
    \frac{\phi 
    \bar{R}^i}{N_i(t-1)} \sum_{k=1}^{\alpha}
    k  B(k) \nonumber \\
    & \qquad +
    \bar{R}^i\sqrt{
    \frac{2\ln (t-1) \sum_{k=1}^{\alpha}
    \left(B(k)\right)^2
    }{N_i(t-1)}
    } \label{eq:c} .
\end{align}
The algorithm then pulls the arm $i$ with the highest upper confidence bound $u_{t-1}^i$ and observes its rewards.
\subsection{Regret Upper Bound of \alg}
\begin{theorem}
\label{Thm:UpperBound}
In the TP-MAB setting with $\beta$-spread reward, the regret of \alg \ after $T$ time steps with B(k) matching the PMF $P_\mathbb{F}(k)$ of random variable $Y$ is upper bounded as

\begin{align*}
&\regret_T(\alg) \leq 
\sum_{i: \mu_i<\mu^*} \frac{4 \ln T\SIGDOLLAR }{\Delta_i} \\
& \quad \qquad \cdot 
\left( 1+ \sqrt{1+ \frac{\Delta_i \phi \SIG}{\bar{R}^i \ln T \sum_{k=1}^{\alpha}
    \left(B(k)\right)^2 }}\right) \\
&  \quad \qquad + 2\phi \SIG \sum_{i: \mu_i<\mu^*} \bar{R}^i +\left(1+\frac{\pi^2}{3}\right) \sum_{i: \mu_i<\mu^*} \Delta_i
\end{align*}
\label{proofFRStep}
\end{theorem}

Observe that $\SIG=\EX[Y]$, meaning the expected value of our random spread variable $Y$ influences the upper bound of the algorithm. Another interesting factor is $\sum_{k=1}^{\alpha}
    \left(B(k)\right)^2$. This factor can be seen as an approximation of the \textit{index of coincidence} \cite{friedman1987index} between rewards. The index of coincidence determines the probability of two reward points being observed in the same $z$-group. Its minimal value equals $\frac{1}{\alpha}$ and occurs when the $\alpha$-smoothness property holds (uniform distribution). The value is maximal and equal to 1 if all rewards fall into one $z$-group. 

\subsubsection{Comparison with the Upper Bound of \texttt{TP-UCB-FR} given in \texorpdfstring{{}\protect\cite{Romano2022}}{[Romano et al., 2022]}}
Let us compare our upper bound given in Theorem \ref{Thm:UpperBound} with the upper bound given in \cite{Romano2022}. For the latter bound to hold, the $\alpha$ estimate given as input to their algorithm has to match the $\alpha$ of the real reward distribution as well.
Note that $\EX[Y]= \frac{\alpha+1}{2}$ in case of $\alpha$-smoothness. For $Y \sim \mathbb{F}$ with $\EX[Y]< \frac{\alpha+1}{2}$ our upper bound on the regret is lower. 
Furthermore, choosing a $\beta$-spread distribution $\mathbb{F}$ with a low mean and a low index of coincidence will result in a better upper bound using Theorem \ref{Thm:UpperBound} compared to choosing $\mathbb{F}$ with rewards centered towards the end (high mean) and not spread out (high index of coincidence). 


\subsubsection{Proof Sketch of Theorem \ref{Thm:UpperBound}}    
Here we provide a proof sketch for Theorem \ref{Thm:UpperBound}. \iftoggle{long-version}{The complete proof can be in Appendix \ref{App:ProofMainThm}}{Please refer to the supplementary material provided with the extended version of the article for the complete proof}.

The approach can be divided into three steps.
Firstly, we show that the probability that an optimal arm is estimated significantly lower than its mean is bounded by $t^{-4}$. Secondly, we show the probability of a suboptimal arm being estimated significantly higher than its mean is bounded by $t^{-4}$. Finally, we evaluate how the algorithm performs in distinguishing the difference between the optimal and suboptimal arms.

\section{Experimental Results}
\label{sec:ExpResults}
In this section, we compare our proposed algorithm \alg \ with \texttt{TP-UCB-FR} \cite{Romano2022}, \texttt{UCB1} \cite{Auer2002}, and \texttt{Delayed-UCB1} \cite{Joulani2013}. 
We observe how well \alg \ performs in settings with different reward distributions. We use the experimental settings proposed by \cite{Romano2022}. That is, two synthetically generated environments and a real-world playlist recommendation scenario. In these settings, we inherit learners used in the provided experiments in \cite{Romano2022}, and create new learner configurations using \alg. As input distributions for the new learners, we use Beta-Binomial distributions with unique parameter values for each learner. The Beta-Binomial distribution gives us the opportunity to model extreme scenarios, which should result in more insightful experimental results. We observe that other distributions do not grant the flexibility of a Beta-Binomial distribution, as demonstrated in experiments deferred to 
\iftoggle{long-version}{Appendix \ref{App:additionalexperiments}.}{the appendix given with the supplementary material for this article.} In the plots under this section, we use the notation \alg$(\alpha, \texttt{dist\_name})$ to denote a learner for our algorithm, where $\texttt{dist\_name}$ is the name of the Beta-Binomial distribution for which the exact parameters are shown in Table \ref{tab:parameters}. Additional details about the used Beta-Binomial distributions and experimental settings are given in 
\iftoggle{long-version}{Appendix \ref{App:additionalexperiments}.}{the appendix given with the supplementary material for this article.} 
\begin{table}
    \centering
    \begin{tabular}{lll}
        \toprule
        Distribution name & $\alpha$ & $\beta$ \\
        \midrule
        \texttt{extreme\_begin}   & 1   & 100 \\
        \texttt{very\_begin}      & 1   & 16  \\
        \texttt{begin}           & 2   & 8   \\
        \texttt{begin\_middle}    & 2   & 4   \\
        \texttt{middle}          & 5   & 5   \\
        \texttt{middle\_end}      & 4   & 2   \\
        \texttt{end}             & 8   & 2   \\
        \texttt{very\_end}        & 16  & 1   \\
        \bottomrule
    \end{tabular}
    \caption{Parameter values for Beta-Binomial distributions}
    \label{tab:parameters}
\end{table}

\subsection*{Setting 1: Uniform Reward Distribution}
In this setting, we evaluate the influence of $\alpha$ on \alg. We set $K = 10$, $\tau_{\max} = 100$ rounds, and the maximum reward such that it is more difficult for a learner to converge to the optimal arm, by letting $\overline{R}^i = 100\zeta^i$ where $\zeta \in \{1, 3, 6, 9, 12, 15, 18, 21, 22, 23\}$. The aggregate rewards are s.t. $Z^{i}_{t,k} \sim \frac{\overline{R}^i}{\alpha} U[0,1]$, and we use a setting $\alpha$-smoothness constant of $\alpha=20$. We run the algorithm over a time horizon $T = 10^5$, and average the results over $100$ independent runs. 
We run the setting for $\alpha_{est} \in \{5, 10, 20, 25, 50\}$, where $\alpha_{est}$ is the estimation of the $\alpha$-smoothness constant. To mimic real-world applications where the underlying data generating distribution and $\alpha$ values are possibly unknown, we use $\alpha_{est}$ to estimate this constant.

\subsubsection*{Results}
\begin{figure}
  \includegraphics[width=\linewidth]{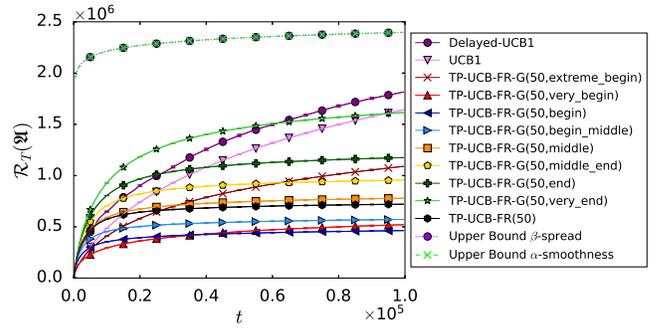}
  \caption{Regret against time for Setting 1 with $\alpha_{est} = 50$}
  \label{fig:setting1alpha50}
\end{figure}

Let us focus on the results for $\alpha_{est} = 50$\footnote{Full results for other values of $\alpha_{est}$ are deferred to 
\iftoggle{long-version}{Appendix \ref{App:Experiments}.}{the appendix given with the supplementary material for this article.}}.
Its performance is plotted against time in Figure \ref{fig:setting1alpha50}, along with the theoretical upper-bounds for the corresponding setting. Note that the upper-bounds for $\beta$-spread and $\alpha$-smoothness are equal, which is expected in this setting with uniform spread. The high upper bound at low values for $t$ is caused by constants in the upper-bound equation, which are dependent on the experimental set-up. The maximum cumulative reward of the arms is the most dominant factor.
First, note that it has been shown by \cite{Romano2022} that optimistic (large) values for $\alpha_{est}$ lead to better performance in practice. However, overly optimistic values for $\alpha_{est}$ do not necessarily lead to better performance. Thus, \texttt{TP-UCB-FR} is largely influenced by the mis-specification of $\alpha_{est}$.
Note that the learners \alg$(50,\texttt{begin})$ and \alg$(50,\texttt{begin\_middle})$ perform significantly better than the \texttt{TP-UCB-FR} learner proposed by \cite{Romano2022}. In fact, \alg$(50,\texttt{begin})$ and \alg$(50,\texttt{begin\_middle})$ are approximately asymptotically parallel to \texttt{TP-UCB-FR}, granting significant performance gains for higher time horizons $T$ as well. This implies that our contribution improves the performance bound in the setting by \cite{Romano2022}.

Further results show that, for lower $\alpha_{est}$ values, \alg \ learners with begin-oriented distributions perform slightly better. In the specific case of $\alpha_{est} = 5$, numerical analysis of the exact regret results deferred to
\iftoggle{long-version}{Appendix \ref{App:Experiments}}{the appendix given with the supplementary material for this article}, shows that our proposed learner \alg$(5, \texttt{begin\_middle})$ performs $\approx 4.5\%$ better than the \texttt{TP-UCB-FR} learner by \cite{Romano2022}. Furthermore, as $\alpha_{est}$ starts to increase, the performance of begin-oriented \alg \ learners increases faster than that of \texttt{TP-UCB-FR}, resulting in an improvement of $\approx 22.1\%$ for $\alpha_{est} = 20$, and to $\approx 36.1\%$ for $\alpha_{est} = 50$. We observe that \alg$(\alpha_{est}, \texttt{begin})$ is essentially the 'ideal' learner, since it always delivers better performance than the learner by \cite{Romano2022} in the tested settings. These results suggest that the issue of being overly/underly optimistic is essentially inherited from the setting by \cite{Romano2022} in a different shape. Overly-optimistic\footnote{We consider a \alg \ learner to be 'optimistic' if it has a 'tail-oriented' Beta-Binomial distribution, and thus expects most of the rewards to be distributed across the first $z$-groups} \alg \ learners perform worse in general. This can be attributed to the fact that overly-optimistic learners generally have an assumed distribution with a high index of coincidence, because the rewards are assumed to be more concentrated at the beginning or at the end. The indices of coincidence for the 'extreme begin' and 'very end' learners in Setting 1 with $\alpha_{est}=50$ are $\approx 0.51$ and $0.14$, respectively. These are significantly higher than $\approx 0.05$, for both the 'begin' and the 'end' learner in the same setting. Similarly, underly-optimistic learners with a middle-oriented distribution also perform poorly. This can be explained by the expected value of their assumed distributions,  which is higher than the begin-oriented distributions as indicated by their name.

\subsection*{Setting 2: Non-Uniform Reward Distributions}
The second setting aims to test the performance of the \alg \ algorithm in scenarios where the distribution of the aggregate reward over the time steps after an arm pull is non-uniform. The distribution of rewards in this setting are s.t. $Z^{i}_{t,k} \sim \frac{\overline{R}^i}{\alpha} Beta[a^i_k,b^i_k]$ where $Beta$ is a Beta distribution with $a, b$ s.t. rewards are distributed according to the spread of the corresponding setting. Again, we model $K=10$ arms, an $\alpha$-smoothness constant of $\alpha=20$ and a maximum reward s.t. convergence to an optimal arm takes longer. That is, $\overline{R}^i = 100\zeta^i$ with $\zeta \in \{1, 3, 6, 9, 12, 15, 18, 21, 22, 23\}$. However, there is a difference in the $\tau_{\max}$, $\alpha_{est}$ and the parameters used for the assumed Beta distribution by the learners. The exact configurations can be found in \iftoggle{long-version}{Appendix \ref{App:Experiments}}{the appendix given with the supplementary material for this article}. In general, there are 12 combinations consisting of 4 configurations with 3 scenarios each. The configurations differ in $\tau_{\max}$ and $\alpha_{est}$, whereas the scenarios differ in distribution parameters. Generally, there is one uniform scenario (equal to Setting 1), one where the rewards are observed late after the pull (Setting 2.1), and one where the results are observed just after the pull (Setting 2.2). We use learners with the same estimated distributions as in Setting 1 (see Table~\ref{tab:parameters}).

\subsubsection*{Results}
Running the proposed Setting 2.1 and 2.2 for $\tau_{\max} = 100$ and $\alpha_{est} = 50$ produces results that are visually identical to the results from Setting 1 given in Figure \ref{fig:setting1alpha50}. However, analysis of the numerical results reveals that differences exist. The cumulative regret generally increases slightly from Setting 2.2 to Setting uniform and then to Setting 2.1.

\begin{figure}
  \includegraphics[width=\linewidth]{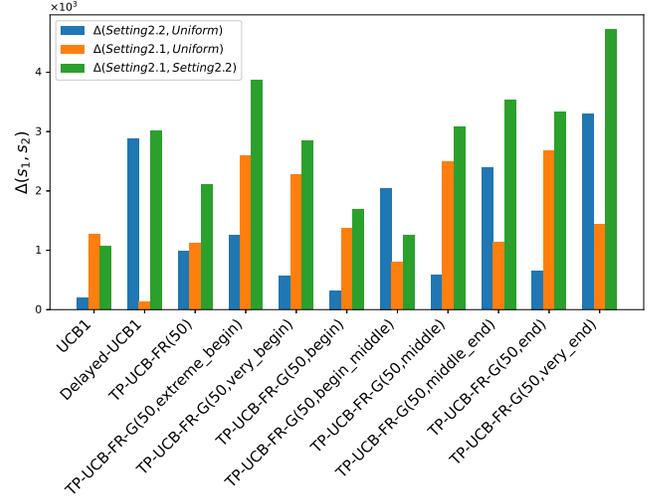}
  \caption{Differences in cumulative regret between scenarios for learners in Setting 2 with $\tau_{\max} = 100, \alpha_{est}=50$}
  \label{fig:diffs_1_3}
\end{figure}

In Figure \ref{fig:diffs_1_3}, we illustrate the differences in cumulative regret between settings where aggregate rewards are distributed non-uniformly. Let us denote $\Delta(s_1, s_2)$ for $s_1, s_2 \in \{\texttt{Setting 2.1}, \texttt{Setting 2.2}, \texttt{Uniform}\}$ as the absolute difference in cumulative regret between Settings $s_1$ and $s_2$. Note that the differences presented in Figure \ref{fig:diffs_1_3} are only marginal. For example, $\Delta(\texttt{Setting 2.1}, \texttt{Setting 2.2}) \approx 4.8 \times 10^3$ for learner \alg$(50, \texttt{very\_end})$ which is the highest difference in average regret observed across all compared settings. Since the regret of \alg$(50, \texttt{very\_end})$ averaged over $T$ is $\approx 1.61 \times 10^6$, the observed change of $\approx 0.3\%$ is neglectable. Furthermore, the same experiment performed with different values for both $\tau_{\max}$ and $\alpha_{est}$ seems to confirm the same marginal change. As an example, Setting 2 for $\tau_{\max} = 200$ and $\alpha_{est} = 20$ results in a maximum change in average regret of only $\approx 0.5\%$.\\
These results suggest that the performance of \alg \ learners in a setting where aggregate rewards are distributed uniformly is indistinguishable from a setting where aggregate rewards are distributed non-uniformly. Therefore, we can align with the conclusion of Setting 1; \alg$(\alpha, \texttt{begin})$ delivers a significant performance increase compared to the learner proposed by \cite{Romano2022}. The gain that we observe for the mentioned settings is as high as $\approx 48.2\%$. An extensive performance summary is deferred to \iftoggle{long-version}{Appendix \ref{App:Experiments}}{the appendix given with the supplementary material for this article}. Again, due to the flexibility of choice for a distribution, there is potential for even higher performance gains.

\begin{figure}
  \includegraphics[width=\linewidth]{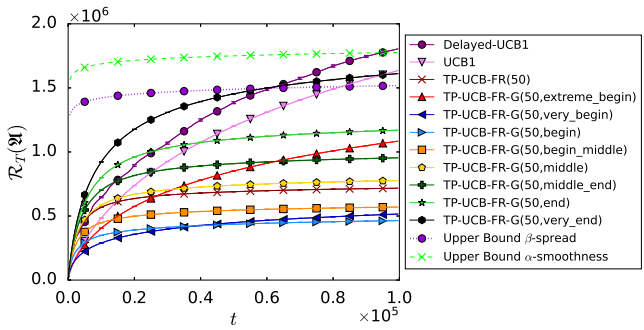}
  \caption{Regret against time for Setting 2.2 with $\tau_{\max} = 100$ and $\alpha_{est} = 50$}
  \label{fig:setting2.2.10050}
\end{figure}

In Figure~\ref{fig:setting2.2.10050}, the theoretical upper bound of \alg\ as well as the the upper bound of the \texttt{TP-UCB-FR} algorithm is plotted on top of the results for the Setting 2.2. 
The figure shows that the upper bound proposed in this article is tighter in this setting. Note that the theoretical upper bounds for \alg\ and \texttt{TP-UCB-FR} only hold for specific learners that assume the data generating distribution precisely and that the 'very end' learner exceeds the $\beta$-spread upper bound. This shows another reason to estimate the assumed distribution optimistically.


\subsection*{User recommendations: Spotify Playlists}
\begin{figure}
  \includegraphics[width=\linewidth]{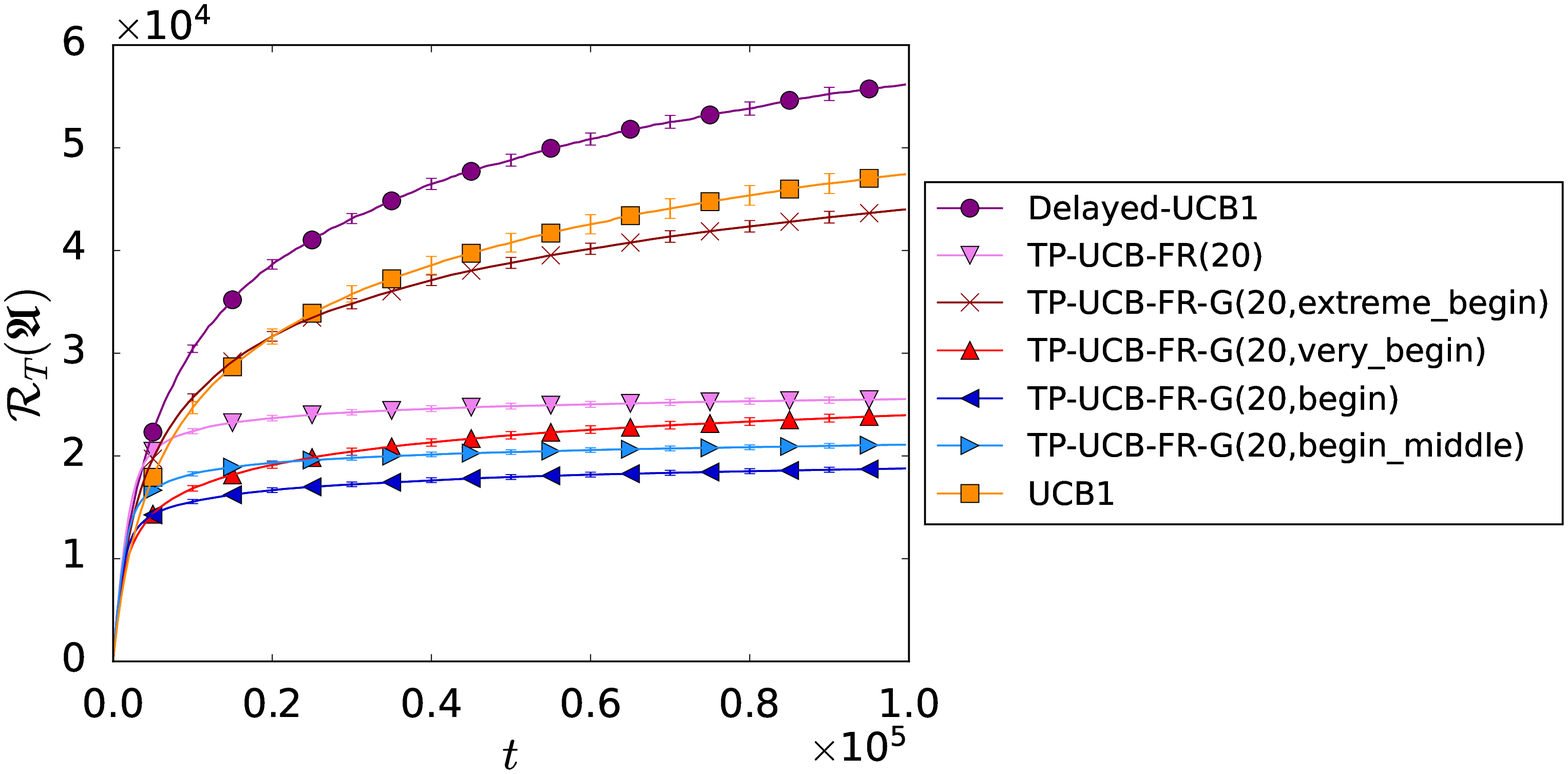}
  \caption{Regret against time for begin-oriented learners in the Spotify Setting}
  \label{fig:settingspotify}
\end{figure}
To verify our algorithm on a real-world dataset, we solve the user recommendation problem introduced by \cite{Romano2022}, using the Spotify dataset introduced in \cite{Brost2019:journals/corr/abs-1901-09851}. We select the $K = 6$ most played playlists as the arms to be recommended. Each time a playlist $i$ is selected, the corresponding reward realizations $x^i_t$ for the first $N = 20$ songs are sampled from the dataset. In this setting, the $\alpha$-smoothness is $\alpha = 20$, the maximum delay $\tau_{\max} = 4N = 80$ and the results are averaged over $100$ independent runs.

\subsubsection*{Results}
In Figure \ref{fig:settingspotify}, we observe that optimistic learners significantly outperform the learner \texttt{TP-UCB-FR}$(20)$ introduced by \cite{Romano2022}. Let us focus on the learner \alg$(20, \texttt{begin})$, since it is by far the best performing learner. We observe that this learner has a decrease of $\approx 26.3\%$ in regret, averaged over the time horizon $T$, when compared to \texttt{TP-UCB-FR}$(20)$. Table \ref{tab:spotifysummary} provides a complete summary of the performance gains of \alg \ learners in the Spotify setting.

\begin{table}
    \centering
    \begin{tabular}{lll}
        \toprule
        Learner & Regret & Decrease \\
        \alg$(\alpha_{est} = 20)$  & ($\times 10^4$) & (\%)     \\
        \midrule
        \texttt{extreme\_begin} & 4.40 &  $\approx -72.5$ \\
        \texttt{very\_begin}    & 2.40 & $\approx 5.9$   \\
        \textbf{\texttt{begin}} & $\boldsymbol{1.88}$ & $\boldsymbol{\approx 26.3}$  \\
        \texttt{begin\_middle}  & 2.11 & $\approx 17.2$  \\
        \bottomrule
    \end{tabular}
    \caption{\alg \ Learners and their decrease in regret compared to the regret of \texttt{TP-UCB-FR}$(20) = 2.55 \times 10^4$}
    \label{tab:spotifysummary}
\end{table}

A second observation we make is that, aligning with the conclusion from Setting 1, overly-optimistic learners such as \alg(20,\texttt{extreme\_begin}) perform significantly worse than \texttt{TP-UCB-FR}$(20)$. As shown in Table \ref{tab:spotifysummary}, the average regret increases by $\approx 72.5\%$. Nevertheless, since \alg$(20, \texttt{begin})$ performs significantly better than \texttt{TP-UCB-FR}$(20)$ for larger $t$, it is even better suited to provide playlist recommendations.

\section{Concluding Remarks and Future Work}
In this article, we model sequential decision-making problems with delayed feedback using a novel formulation called multi-armed bandits with generalized temporally-partitioned rewards. To generalize delayed reward distributions, we introduce the $\beta$-spread property. 
A lower bound for this TP-MAB setting with the $\beta$-spread property is provided that can be tighter than the lower bound for the TP-MAB setting with the $\alpha$-smoothness property. More specifically, we show that even when $\alpha$ is equal for both settings, a high mean or high index of coincidence for the assumed distribution leads to a tighter bound in the setting introduced in this article. We also introduce the \alg \ algorithm 
that exploits the $\beta$-spread property and show that in certain scenarios the upper bound of this algorithm can be lower than the upper bound of the \texttt{TP-UCB-FR} algorithm, and therefore also lower than the upper bounds of the classical UCB1 and Delayed-UCB1 algorithms.
%
Finally, we empirically show that our algorithm outperforms \texttt{TP-UCB-FR} and other UCB algorithms over a wide range of experiments with synthetic and real-world data. The decrease in cumulative regret reaches 26.3\% against the state of the art \texttt{TP-UCB-FR} algorithm in the experiment with real-world data.

Directions for future work include:
\begin{enumerate}
    \item Currently, the $\beta$-spread property is restricted to discrete probability distributions bounded by a finite domain of size $\alpha$. An interesting future research topic could be to look into the removal of these restrictions, as this could lead to more flexibility in our algorithm, and broaden its practical applications.
    \item Another possible extension could be to consider settings in which the arms are treated as subsets, such that each subset of arms is assigned different $\alpha$-values and distributions. This could be advantageous in settings where arms are treated as clusters e.g., the setting by \cite{pandey2007multi}
    \item Moreover, another interesting topic is to look at scenarios where the time span over which the reward is partitioned i.e., $\tau_{\max}$, is variable. In our current work, we assume it to be fixed and consistent across arms. Removing this assumption could be beneficial if we want to consider practical applications where $\tau_{\max}$ is not fixed, such as observing the reward of an online advertisement (clicks on the advertisement) over the lifetime ($\tau_{\max}$) of that advertisement, which is variable.
\end{enumerate}



\bibliographystyle{named}
\bibliography{references}

\clearpage

\appendix
\section{Proof of Theorem \ref{Thm:LowerBound}}
\label{App:Proof:Thm:LowerBound}
\begin{proof}
    The proof follows along the lines of the proof of Theorem 2.2 from \cite{Bubeck-Bianchi2012}, which is based on \cite{LAI19854}. 
Because the $\beta$-spread property has no effect on the cardinality $\phi$ of the $z$-groups, we can generalize to a setting where multiple rewards are earned by a single arm pull.
Let us define an auxiliary TP-MAB setting in which:
\begin{itemize}
    \item only two arms exist with expected values $\mu_1$ and $\mu_2$ s.t. $\mu_2 < \mu_1 < 1$.
    \item upper bound on the reward for each arm is equal to the maximum upper bound, i.e., $\overline{R}^i_t = \overline{R}_{\max}$
    \item The total rewards in each $z$-group, $Z^i_{t,k}$, are independent, and the expected value of the rewards in each $z$-group is $\mathrm{B}(k) \cdot \mu_i$.
    \item The total reward in $z$-group, $Z^i_{t,k}$, is a scaled Bernoulli random variable s.t. $Z^i_{t,k}\in\{0, \mathrm{B}(k) \cdot \overline{R}_{\max}\}$
    \item Pulling an arm at time $t$ provides rewards $\{Z^i_{t,1}, ..., Z^i_{t,\alpha}\}$ that can all be observed immediately at the time of the pull.
\end{itemize}
In this proof of the lower bound, we trivially observe that finding the optimal arm in a setting in which all of the partial rewards are observed at once can never be more difficult than in a setting in which rewards are spread out over a set of rounds $\{t, t+1, ..., \tau_{max}\}$. As a result, a lower bound in this defined setting corresponds to a lower bound in our $\beta$-spread setting.

To give an idea of how good an arm is compared to its maximum, we derive a new alternative mean for each arm as $\mu_{A_i}=\frac{\mu_i}{\overline{R}_{\max}}$. Note that $\mu_{A_i} < 1$ as mentioned before. 

Let $\mathbb{E}[\mathrm{N}_i(T)]$ denote the \textit{expected} number of times an arm $i$ is pulled over a set of rounds $T$. To compute $\mathbb{E}[\mathrm{N}_1(\{t, t+1, ..., \tau_{max}\})]$ and $\mathbb{E}[\mathrm{N}_2(\{t, t+1, ..., \tau_{max}\})]$, we can use the scaled reward values without loss of validity.
If we now consider a second, modified instance of the above TP-MAB setting, with the only difference being that arm 2 is now the optimal arm s.t. $\mu_{A_1} < \mu'_{A_2} < 1$, we can show that the learning agent choosing the arms cannot distinguish between the different instances. This reasoning implies a lower bound on the number of times a suboptimal arm is played. We know that $x \mapsto \KLdiv(\mu_{A_1}, x)$ is a continuous function, and we can find a $\mu'_{A_2}$ for each $\epsilon > 0$, such that:
\begin{equation}
    \label{eq:klinequality}
    \KLdiv(\mu_{A_2}, \mu'_{A_2}) \leq (1+\epsilon) \KLdiv(\mu_{A_2}, \mu_{A_1})
\end{equation}
The proof follows the steps given in the work by \cite{Bubeck-Bianchi2012} to derive a lower bound for any uniform policy $\mathfrak{U}$.
\subsection*{Step 1: $\mathbb{P}(C_t) = o(1)$}
For this proof, we change the notation of the rewards slightly such that each variable in the sequence $Z^i_{1, 1}, ..., Z^i_{n, \alpha}$ represents the cumulative reward of an arm $i$ when pulled $n$ times, at timestep $k \in \{1, 2, ..., \alpha\}$. $Z^i_{s, k}$ for $s \in \{1, 2,...,n\}$ represents the cumulative reward of an arm $i$ after the $s$'th pull at timestep $k$ after a pull. Using this notation, we can define the empirical estimate of $\KLdiv(\mu_{Z_2}, \mu'_{Z_2})$ as:

\begin{equation*}
    \widehat{\KLdiv}_{\alpha \beta} := \sum_{n=1}^s\sum_{k=1}^\alpha \mathrm{ln} \frac{\mu_{A_2}Z^2_{n, k} + (1-\mu_{A_2})(1-Z^2_{n, k})}{\mu'_{A_2}Z^2_{n, k} + (1-\mu'_{A_2})(1-Z^2_{n, k})}
\end{equation*}

Using this, we define an event that links the behavior of the original agent to the modified version

\begin{equation}
    C_t \defined \left\{\alpha N_2(t)<f_t \quad \text { and } \quad \widehat{\KLdiv}_{\alpha N_2(t)} \leq(1-\varepsilon / 2) \ln t\right\}
    \label{eq:defCt}
\end{equation}

with 

\begin{equation*}
    f_t = \left(\frac{2}{\alpha + 1}\SIG \cdot\alpha\INDEXCOINCIDENCE  \right)\frac{1-\epsilon}{\KLdiv(\mu_{A_2}, \mu'_{A_2})} \ln t
\end{equation*}

Using the change of measure identity defined in \cite{Bubeck-Bianchi2012} and the second inequality in the definition of $C_t$ given in Eq. \eqref{eq:defCt}:

\begin{equation*}
    \mathbb{P}^{\prime}\left(C_t\right)=\mathbb{E}\left[1_{C_t} \exp \left(-\widehat{\KLdiv}_{\alpha N_2(t)} \right)\right] \geq e^{-(1-\varepsilon / 2) \ln t} \mathbb{P}\left(C_t\right)
\end{equation*}

Then, we first rearrange the terms of the above inequality to obtain

\begin{align*}
    \mathbb{P}\left(C_t\right) &\leq t^{(1-\varepsilon / 2)} \mathbb{P}^{\prime}\left(C_t\right) \\
    &\leq t^{(1-\varepsilon / 2)} \mathbb{P}^{\prime}\left(\alpha N_2(t)<f_t\right) \\
    &\leq t^{(1-\varepsilon / 2)} \frac{\mathbb{E}^{\prime}\left[t-N_2(t)\right]}{t-f_t / \alpha} \\
    &=o(1)
\end{align*}   

In the equations above, we use $\mathbb{P}^{\prime}\left(C_t\right) \leq \mathbb{P}^{\prime}\left(\alpha N_2(t)<f_t\right)$, Markov's inequality and the fact that the policy $\mathfrak{U}$ is uniformly efficient (i.e. $\EX[N_2(t)] = o(t^{\gamma})$ with $\gamma<1$).

\subsection*{Step 2: $\mathbb{P}\left(\alpha N_2(t)\leq f_t\right)=o(1)$}
 Using Theorem 2.2 from \cite{Bubeck-Bianchi2012} and observing that we always have
\begin{enumerate}
    \item $\SIG \geq 1 \implies \frac{2}{\alpha + 1}\SIG \geq \frac{2}{\alpha + 1}$
    \item $\INDEXCOINCIDENCE \in [\frac{1}{\alpha},1]\implies \alpha\INDEXCOINCIDENCE \in [1,\alpha]$
    \item $\frac{2}{\alpha + 1}\SIG\cdot\alpha\INDEXCOINCIDENCE \geq \frac{2}{(\alpha+1)} > 0$
\end{enumerate}

Next, we define two events
\begin{align*}
    E_1 = \alpha N_2(t)<f_t
\end{align*}

and

\begin{align*}
    E_2 &= \biggl( \frac{\alpha + 1}{2\alpha} \cdot \frac{1}{\SIG \cdot \INDEXCOINCIDENCE} \cdot \\
    & \frac{\KLdiv\left(\mu_{Z_2}, \mu_{Z_2}^{\prime}\right)}{(1-\varepsilon) \ln t} \cdot \max_{\beta<f_t / \alpha} \widehat{\KLdiv}_{\alpha \beta} \\
    & \leq \frac{1-\varepsilon / 2}{1-\varepsilon} \cdot 
      \frac{\alpha + 1}{2\alpha} \cdot \frac{\KLdiv (\mu_{Z_2}, \mu_{Z_2}^{\prime})}{\SIG \cdot \INDEXCOINCIDENCE} \biggr)
\end{align*}

such that we obtain:

\begin{align*}
    o(1) &= \mathbb{P}\left(C_t\right) \leq \mathbb{P}(E_{1} \wedge E_{2})
\end{align*}

    %

Using the strong law of large numbers for the event $E_2$ s.t. $\lim _{t \rightarrow+\infty} \mathbb{P}\left(E_2\right)=1$, we can conclude that $\mathbb{P}\left(E_1\right)=\mathbb{P}\left(\alpha N_2(t)<f_t\right)=o(1)$, and that for $t \rightarrow+\infty$ we have $\mathbb{E}\left[N_2(t)\right]>f_t / \alpha$.

\subsection*{Final Step}
Using Equation (\ref{eq:klinequality}) we know that, for $t \rightarrow+\infty$ :

\begin{align*}
    \mathbb{E}\left[N_2(t)\right] & >f_t / \alpha \\
    &= \frac{2}{\alpha + 1}\SIG\cdot\alpha\INDEXCOINCIDENCE \\
& \qquad \frac{1-\varepsilon}{\alpha \KLdiv\left(\mu_{A_2}, \mu_{A_2}^{\prime}\right)} \ln t \\
& \geq \frac{2}{\alpha + 1}\SIG
 \cdot \alpha\INDEXCOINCIDENCE \\
 & \qquad\frac{1-\varepsilon}{\alpha(1+\varepsilon) \KLdiv\left(\mu_{A_2}, \mu_{A_1}\right)}
\ln t 
\end{align*}

where the theorem statement is derived from the arbitrariness of the value of $\varepsilon$, substituting $\mu_{A_1}$ with $\frac{\mu^*}{\bar{R}_{\max }}$ and $\mu_{A_2}$ with $\frac{\mu_2}{\bar{R}_{\max }}$, and summing over all the sub-optimal arms.
\end{proof}

\section{Proof of Theorem~\ref{Thm:UpperBound}}\label{App:ProofMainThm}


\subsection{Preliminaries}
The relation between the expected amount of sub-optimal arm pulls and the regret of the algorithm is given by

\begin{equation*}
   \regret_T(\alg)=\sum_{i: \mu_i<\mu^*} \Delta_i \mathbb{E}\left[N_i(T)\right]
\end{equation*}
 Let us define the true empirical mean of the cumulative reward of arm $i$ computed over $N_i(t)$ arm pulls:

\begin{equation*}
    \hat{R}_t^{i, \text { true }}:=\frac{1}{N_i(t)} \sum_{h=1}^t r_h^i \ind_{\left\{i_h=i\right\}}
\end{equation*}

The value above assumes that the cumulative reward of an arm pull is known, even if partial rewards are still to come in the future. We bound the difference between the true empirical mean and the observed empirical mean as follows:

\begin{align}
&\hat{R}_t^{i, \text { true }}-\hat{R}_t^i \nonumber \\
&=\frac{1}{N_i(t)} \sum_{h=1}^t \sum_{j=1}^{\tau_{\max }}\left(x_{h, j}^i-\tilde{x}_{h, j}^i\right) \ind_{\left\{i_h=i\right\}} \nonumber \\
&\leq \frac{1}{N_i(t)} \sum_{h=1}^t \sum_{j=1}^{\tau_{\max }}(x_{h, j}^i-\tilde{x}_{h, j}^i) \nonumber \\
&=\frac{1}{N_i(t)} \sum_{h=\max \left\{1, t-\tau_{\max }+2\right\}}^t \sum_{j=t-h+2}^{\tau_{\max }} x_{h, j}^i \label{difftrue} \\
%
&\leq \frac{1}{N_i(t)} \sum_{k=1}^{\alpha}
k \phi \bar{R}^i  B(k)
\label{ineqbeta} \\
& = \frac{\phi \bar{R}^i}{N_i(t)} \sum_{k=1}^{\alpha}
k  B(k)
\end{align}

Eq. \eqref{difftrue} states that the difference between the true and observed mean equals the sum of all future rewards that are yet to be observed for a maximum of $\tau_{\max}-1$ arms that have been pulled. The closer index $h$ gets to the current time $t$, the more pulled arms exist with unobserved rewards. Therefore, the amount of reward that is unobserved can be bounded by looping over all $z$-groups in Eq. \eqref{ineqbeta} to calculate the maximum reward still to be observed and giving higher weight to late $z$-groups through index $k$. Furthermore, Eq. \eqref{ineqbeta} holds because of the $\beta$-spread property.

\begin{fact}[Hoeffding inequality \cite{Hoeffding}]
    Let $X_1, \ldots, X_n$ be random variables in $[0,1]$ such that $\mathbb{E}\left[X_t \mid X_1, \ldots, X_{t-1}\right]=\mu$. Let $S_n=X_1+\cdots+X_n$. Then, for all $a \geq 0$
    
\begin{equation*}
    \quad \mathbb{P}\left\{S_n \leq n \mu-a\right\} \leq e^{-2 a^2/n}.
    \label{hoeffdingboundMainPaper}
\end{equation*}
\end{fact}

\subsection*{Deriving the upper bound} 
By construction of the algorithm \alg, the upper bound on the expected number of times a suboptimal arm $i$ is pulled can be expressed as follows:
\begin{equation}
    \mathbb{E}\left[N_i(t)\right] \leq \ell+\sum_{t=1}^{\infty} \sum_{s=1}^{t-1} \sum_{s_i=\ell}^{t-1} \mathbb{P}\left\{\left(\hat{R}_{t, s}^*+c_{t, s}^*\right) \leq\left(\hat{R}_{t, s_i}^i+c_{t, s_i}^i\right)\right\}
    \label{ineqAuer}
\end{equation}
where $\hat{R}_{t, s}^*$ and $c_{t, s}^*$ are the empirical mean and the confidence term of the optimal arm and $\hat{R}_{t, s_i}^i$ and $c_{t, s_i}^i$ denote the empirical mean and the confidence term for arm $i$.

For (\ref{ineqAuer}) to hold, one of the following three inequalities have to hold as well:
\begin{equation}
\hat{R}_{t, s}^* \leq \mu^*-c_{t, s}^*
\label{optimaltoolow}
\end{equation}
\begin{equation}
    \hat{R}_{t, s_i}^i \geq \mu_i+c_{t, s_i}^i
\label{badtoohigh}
\end{equation}
\begin{equation}
    \mu^*<\mu_i+2 c_{t, s_i}^i
\label{badoptimalequal}
\end{equation}
Let us pay attention to (\ref{optimaltoolow}) first and find the following:

\begin{align*}
     &\mathbb{P}\left(\hat{R}_{t, s}^*-\mu^* 
     \leq-c_{t, s}^*\right) \\
     &=\mathbb{P}\left(\hat{R}_{t, s}^{* \text { true }}-\mu^* \leq-c_{t, s}^*+
    \hat{R}_{t, s}^{* \text { true }}-\hat{R}_{t, s}^*\right) \\
    &\leq  \mathbb{P}\left(\hat{R}_{t, s}^{* \text { true }}-\mu^* \leq-c_{t, s}^*+
    \frac{\phi 
    \bar{R}^i}{s} \sum_{k=1}^{\alpha}
    k  B(k)
    \right) \\
    &= \mathbb{P}\left(s\hat{R}_{t, s}^{* \text { true }} \leq
    s\mu^*-s\sqrt{
    \frac{2\ln t\sum_{k=1}^{\alpha}
    \left(\bar{R}^*B(k)\right)^2
    }{s}
    }\right) \\
    &\leq \exp \left\{-\frac{
    \left(2\sqrt{
    \frac{2\ln t\sum_{k=1}^{\alpha}
    \left(\bar{R}^*B(k)\right)^2
    }{s}
    }\right)
    ^2 s^2}{
    \sum_{l=1}^{s}
    \sum_{k=1}^{\alpha}
    \left(\bar{R}^*B(k)\right)^2}\right\} \\
     &\leq e^{-4 \ln t} \\
     & = t^{-4}
\end{align*}
where we use Hoeffding's inequality (defined in Fact~\ref{hoeffdingboundMainPaper}), the penultimate step and $c_{t, s_i}^i \defined 
    \frac{\phi 
    \bar{R}^i}{s} \sum_{k=1}^{\alpha}
    k  B(k)
    +
    \bar{R}^i\sqrt{
    \frac{2\ln t\sum_{k=1}^{\alpha}
    \left(B(k)\right)^2
    }{s}
    }
$

Similarly, the bound from (\ref{badtoohigh}) can be derived:

\begin{align*}
    \mathbb{P}\left(\hat{R}_{t, s_i}^i-\mu_i \geq c_{t, s_i}^i\right) &\leq \mathbb{P}\left(\hat{R}_{t, s}^{i, \text { true }}-\mu_i \geq \bar{R}^i \sqrt{\frac{2 \ln t}{\alpha s_i}}\right) \\
    &\leq e^{-4 \ln t} \\
    &=t^{-4}
\end{align*}
where we use Hoeffding's inequality (defined in Fact~\ref{hoeffdingboundMainPaper}) and the fact that by definition $\hat{R}_{t, s_i}^{i} \leq \hat{R}_{t, s_i}^{i,true}$. All that is left to do is to consider Eq. \eqref{badoptimalequal}. Let us assume that Eq. \eqref{badoptimalequal} does not hold i.e., $\mu^* \geq \mu^i + 2c^i_{t, s}$. This is equivalent to 

\begin{align*}
    &\Delta_i \geq 2\left(
    \frac{\phi 
    \bar{R}^i}{s_i} \sum_{k=1}^{\alpha}
    k  B(k)
    +
    \sqrt{
    \frac{2\ln t\sum_{k=1}^{\alpha}
    \left(\bar{R}^* B(k)\right)^2
    }{s_i}
    }\right)
%
\end{align*}
Rearranging the terms. 

\begin{align*}
 &\frac{\Delta_i^2}{4} + \frac{\phi^2 
    (\bar{R}^i)^2}{s_i^2} \left(\sum_{k=1}^{\alpha}
    k  B(k)\right)^2
    -
    2\left(\frac{ \Delta_i \phi 
    (\bar{R}^i)}{2 s_i} \left(\sum_{k=1}^{\alpha}
    k  B(k)\right)   \right) \\
    & \geq \frac{2\ln t\sum_{k=1}^{\alpha}
    \left(\bar{R}^*B(k)\right)^2
    }{s_i} \\
& s_i^2\frac{\Delta_i^2}{4} + \phi^2 
    (\bar{R}^i)^2 \left(\sum_{k=1}^{\alpha}
    k  B(k)\right)^2
    -
    2 s_i\bigg(\frac{ \Delta_i \phi
    (\bar{R}^i)}{2     } \left(\sum_{k=1}^{\alpha}
    k  B(k)\right)   \\
    & \qquad + 
    \ln t\sum_{k=1}^{\alpha}
    \left(\bar{R}^* B(k)\right)^2
    \bigg)  \\
    & \geq 0
\end{align*}

By solving for $s_i$, the following can be established:

\begin{align*}
    %
%
s_i &\geq 
    \frac{2 \phi \bar{R}^i \SIG}{\Delta_i}
   +
    \frac{4 \ln t \SIGDOLLAR }{\Delta_i^2}
    \\
    &\qquad +  4 \frac{\sqrt{\left(\ln t \SIGDOLLAR \right) \left( 1 + \frac{\SIG \Delta \phi \bar{R}^i }{\ln t \SIGDOLLAR}\right)}}{\Delta_i^2} \\
  s_i &\geq   
\frac{2\phi \bar{R}^i \SIG}{\Delta_i} + \left( \frac{4 \ln t\SIGDOLLAR }{\Delta^2_i} \right) \\
& \quad \cdot \left( 1+ \sqrt{1+ \frac{\Delta_i \phi \bar{R}^i \SIG}{\ln t \SIGDOLLAR }}\right)
\end{align*}

As a result, we pick 
\begin{align*}
    l &= \Biggl \lceil \frac{2\phi \bar{R}^i \SIG}{\Delta_i} + \left( \frac{4 \ln t\SIGDOLLAR }{\Delta^2_i} \right)\\
    & \qquad \cdot \left( 1+ \sqrt{1+ \frac{\Delta_i \phi \bar{R}^i \SIG}{\ln t \SIGDOLLAR }}\right)\Biggr \rceil
\end{align*}
%
to ensure that the inequality in Eq. \eqref{badoptimalequal} is always false for $s_i \geq l$.
\begin{align*}
     & \mathbb{E}\left[N_i(t)\right] \\
     & \leq \Biggl \lceil \frac{2\phi \bar{R}^i \SIG}{\Delta_i} + \left( \frac{4 \ln t\SIGDOLLAR }{\Delta^2_i} \right)\\
    & \qquad \cdot \left( 1+ \sqrt{1+ \frac{\Delta_i \phi \bar{R}^i \SIG}{\ln t \SIGDOLLAR }}\right)\Biggr \rceil \\
    & \qquad +  \sum_{t=1}^{\infty} \sum_{s=1}^{t-1} \sum_{s_i=\ell}^{t-1}\left[\mathbb{P}\left(\hat{R}_{t, s}^*-\mu^* \leq-c_{t, s}^*\right) \right] \\
    & \qquad + \sum_{t=1}^{\infty} \sum_{s=1}^{t-1} \sum_{s_i=\ell}^{t-1} \left[ \mathbb{P}\left(\hat{R}_{t, s_i}^i-\mu_i \geq c_{t, s_i}^i\right)\right] \\
    &\leq \frac{2\phi \bar{R}^i \SIG}{\Delta_i} + \left( \frac{4 \ln t\SIGDOLLAR }{\Delta^2_i} \right)\\
    & \qquad \cdot \left( 1+ \sqrt{1+ \frac{\Delta_i \phi \bar{R}^i \SIG}{\ln t \SIGDOLLAR }}\right) \\
    & \qquad +1+\sum_{t=1}^{\infty} \sum_{s=1}^{t-1} \sum_{s_i=\ell}^{t-1} 2 t^{-4} \\
     &\leq \frac{2\phi \bar{R}^i \SIG}{\Delta_i} + \left( \frac{4 \ln t\SIGDOLLAR }{\Delta^2_i} \right)\\
    & \qquad \cdot \left( 1+ \sqrt{1+ \frac{\Delta_i \phi \bar{R}^i \SIG}{\ln t \SIGDOLLAR }}\right) \\
    & \qquad + 1 + \frac{\pi^2}{3}
\end{align*}
The theorem statement follows by the fact that $\mathcal{R}_T\left(\mathfrak{U}_{\mathrm{FR}}\right)=\sum_{i: \mu_i<\mu^*} \Delta_i \mathbb{E}\left[N_i(T)\right]$.

\clearpage

\section{Experimental Environment Details}\label{App:Experiments}
\subsection{Technical Details}\label{App:Technicaldetails}
The code has been executed on a server configured with 2 Intel Xeon 4110 \@ 2.1Ghz (32 hyperthreads) CPU's and 384GB RAM. We did not make use of GPU acceleration during the simulation process. The operating system used is Ubuntu 16.04.7 LTS. The code for the simulations is created in Python with version 3.9.12. Furthermore, we use Conda for library management, and for the experiments, the following libraries are used:
\begin{itemize}
    \item numpy 1.23.4
    \item pandas 1.5.1
    \item tqdm 4.64.1
    \item scipy 1.9.3
    \item matplotlib 3.5.3
\end{itemize}
Because this research is partially based on the findings by \cite{Romano2022}, we used their code as a base and made adaptations to run the experiments for our learners\footnote{The code is provided with the supplementary material.}. Overall, running all settings takes approximately 96 hours on the hardware mentioned above.

\subsection{\alg \ learner configurations}\label{App:LearnerBBConfig}
\begin{figure}
  \includegraphics[width=\linewidth]{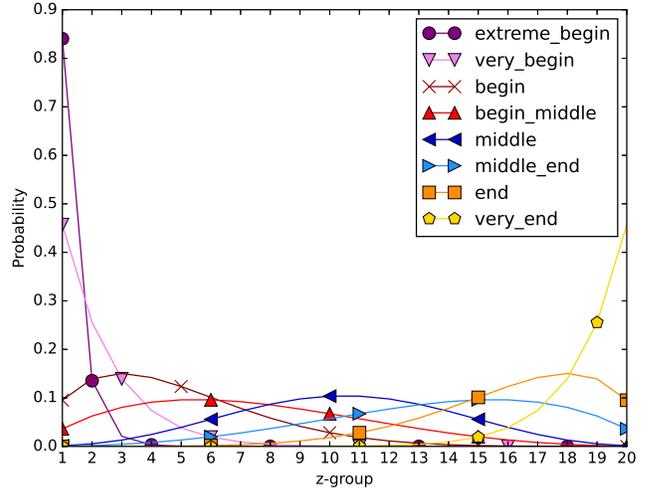}
  \caption{Probability mass functions for different Beta-Binomial configurations}
  \label{fig:parameters}
\end{figure}

We introduce new learners to the experimental settings using a variant of the Beta-Binomial distribution. A brief introduction to the Beta-Binomial distribution itself will be provided first, followed by the introduction of our variation.
 For a more complete overview of Beta-Binomial distributions, we refer to \cite{lee_2012}.\\
Let $N \in \mathbb{N}$, $\alpha \in \mathbb{R}^+$ and $\beta \in \mathbb{R}^+$, and note that $\alpha$ in this setting is some arbitrary constant instead of a spread parameter as it is in the rest of this article. For $x \in \{0,..,N\}$, the probability mass function is then defined as
\begin{equation}
    BetaBinom(N, \alpha, \beta)(x) = \binom{N}{x}\frac{B(x+\alpha, N-x+\beta)}{B(\alpha, \beta)}
\end{equation}
where the $B(u,v)$ is the \textit{beta function} for some $u,v \in \mathbb{R}^+$. As can be seen, the domain of the distribution is not equal to the $\alpha$-sized domain of $\{1,2,\dots, \alpha\}$ required for the $\beta$-spread property to hold. Therefore, we create a variant of the Beta-Binomial distribution that is in this domain in Equation~\ref{eq:betabinomvariation}.

\begin{equation}
\label{eq:betabinomvariation}
    BetaBinomVariant(x) = BetaBinom(\alpha-1, a, b)(x-1)
\end{equation}
where $\alpha$ denotes the amount of $z$-groups and a and b are the distribution parameters. The domain of this function is the desired interval of all integers $\{1,2,\dots, \alpha\}$. Note that formulas for calculating the mean or variance do not hold anymore for this variant.
 For each experimental setting, we add 8 synthetic learners that are distributed according to this variant. Their exact parameters are given in Table \ref{tab:parameters}, and the corresponding probability mass functions are also plotted in Figure \ref{fig:parameters}. Note that the distribution names are chosen according to their 'center' on the $x$--axis. 
That is, the location on the $x$-axis with the highest observed probability.

\subsection{Experimental Settings}\label{App:Experimentalsettings}
In this section, we detail the experiment settings presented in Section \ref{sec:ExpResults} and some further experiments that have been run to confirm the results in the main article.

\subsubsection{Setting 1}
\begin{figure}
  \includegraphics[width=\linewidth]{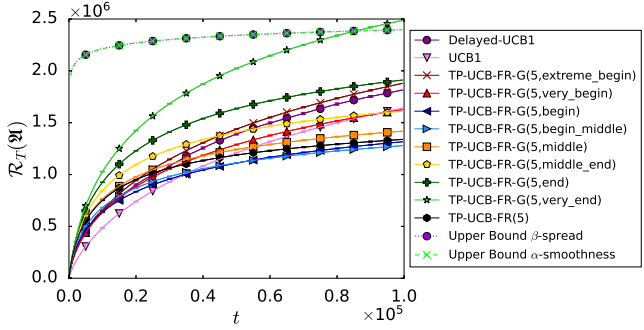}
  \caption{Regret against time for Setting 1 with $\alpha_{est} = 5$}
  \label{fig:setting1alpha5}
\end{figure}

\begin{figure}
  \includegraphics[width=\linewidth]{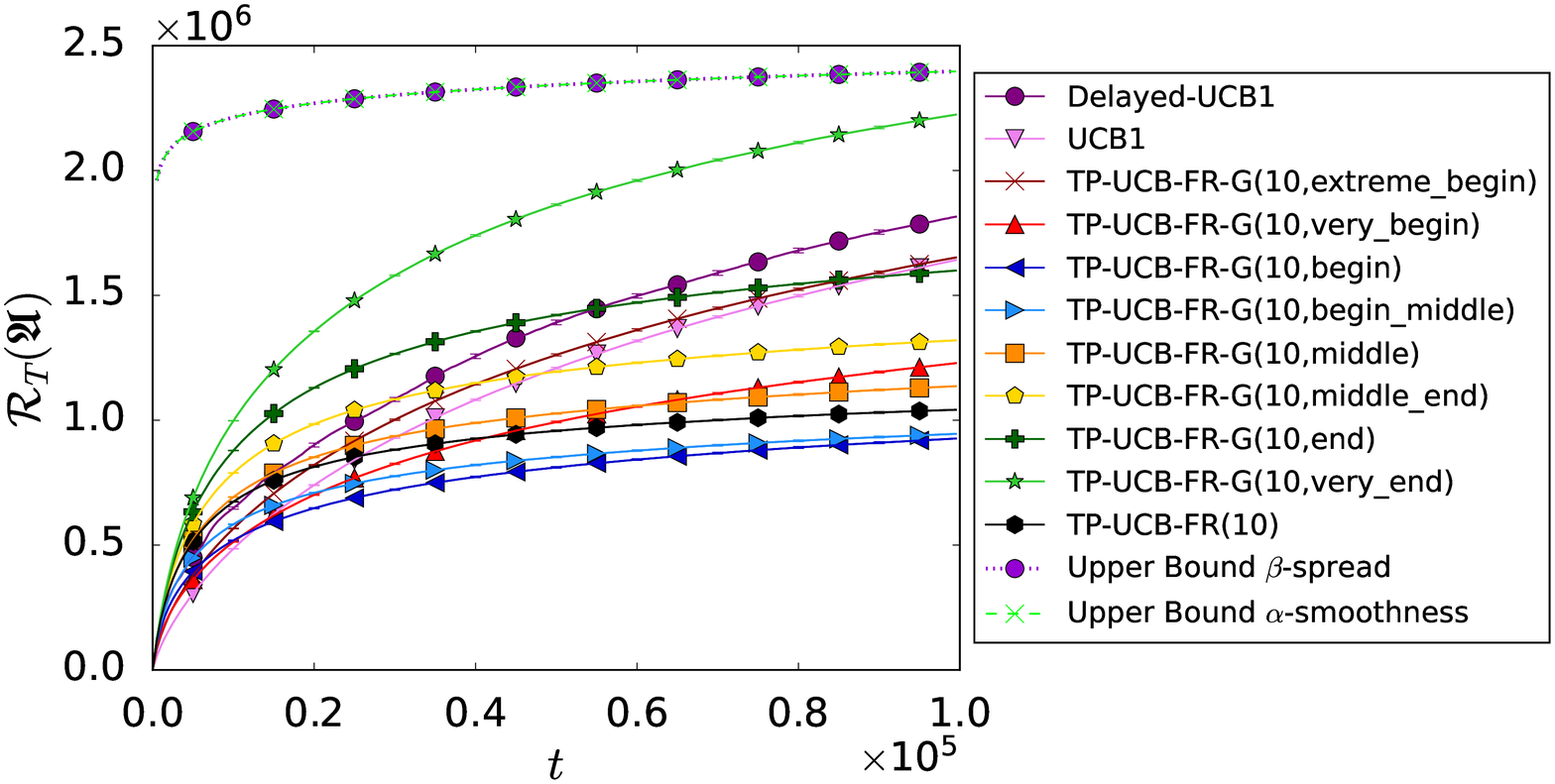}
  \caption{Regret against time for Setting 1 with $\alpha_{est} = 10$}
  \label{fig:setting1alpha10}
\end{figure}

\begin{figure}
  \includegraphics[width=\linewidth]{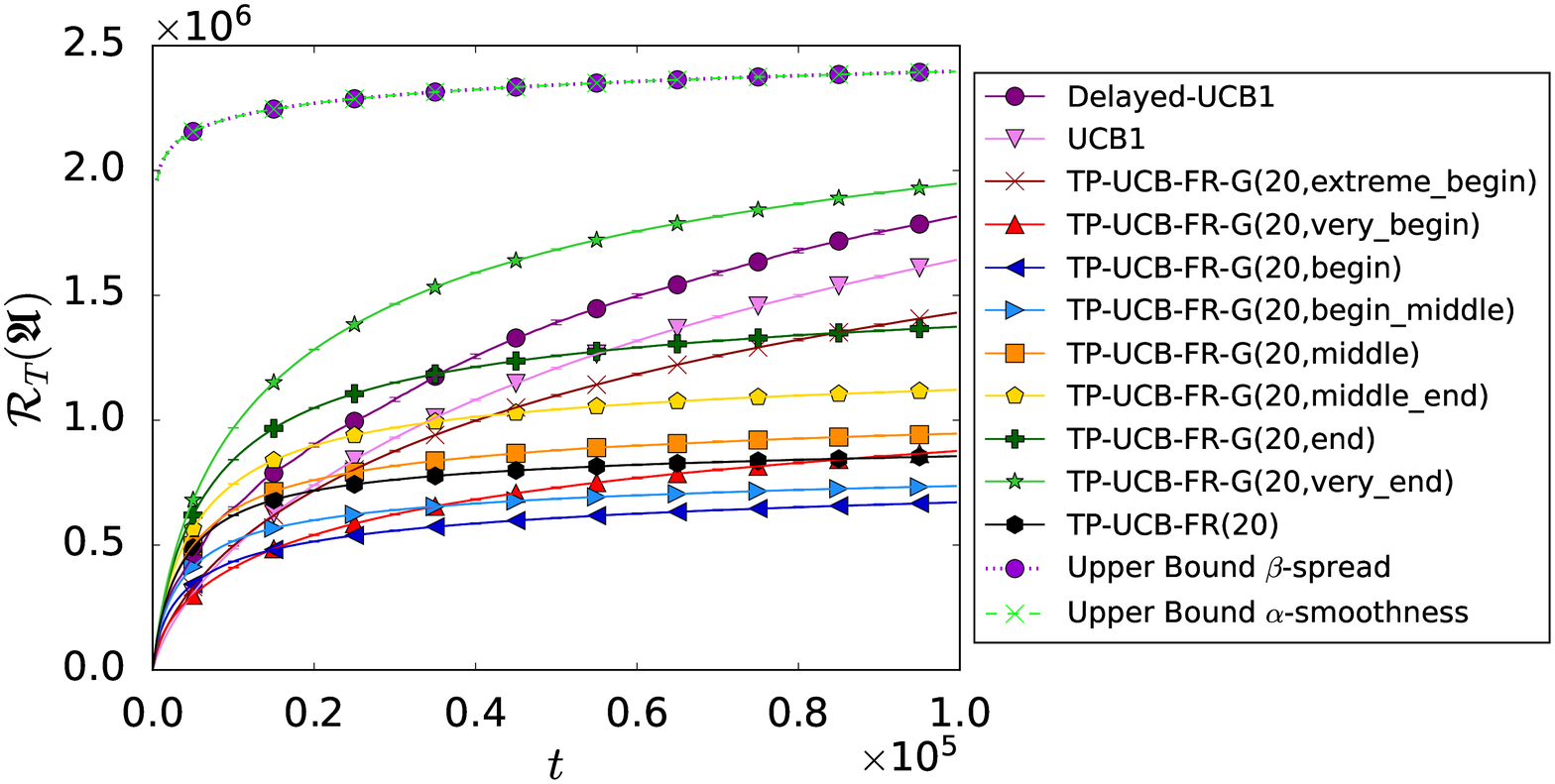}
  \caption{Regret against time for Setting 1 with $\alpha_{est} = 20$}
  \label{fig:setting1alpha20}
\end{figure}

\begin{figure}
  \includegraphics[width=\linewidth]{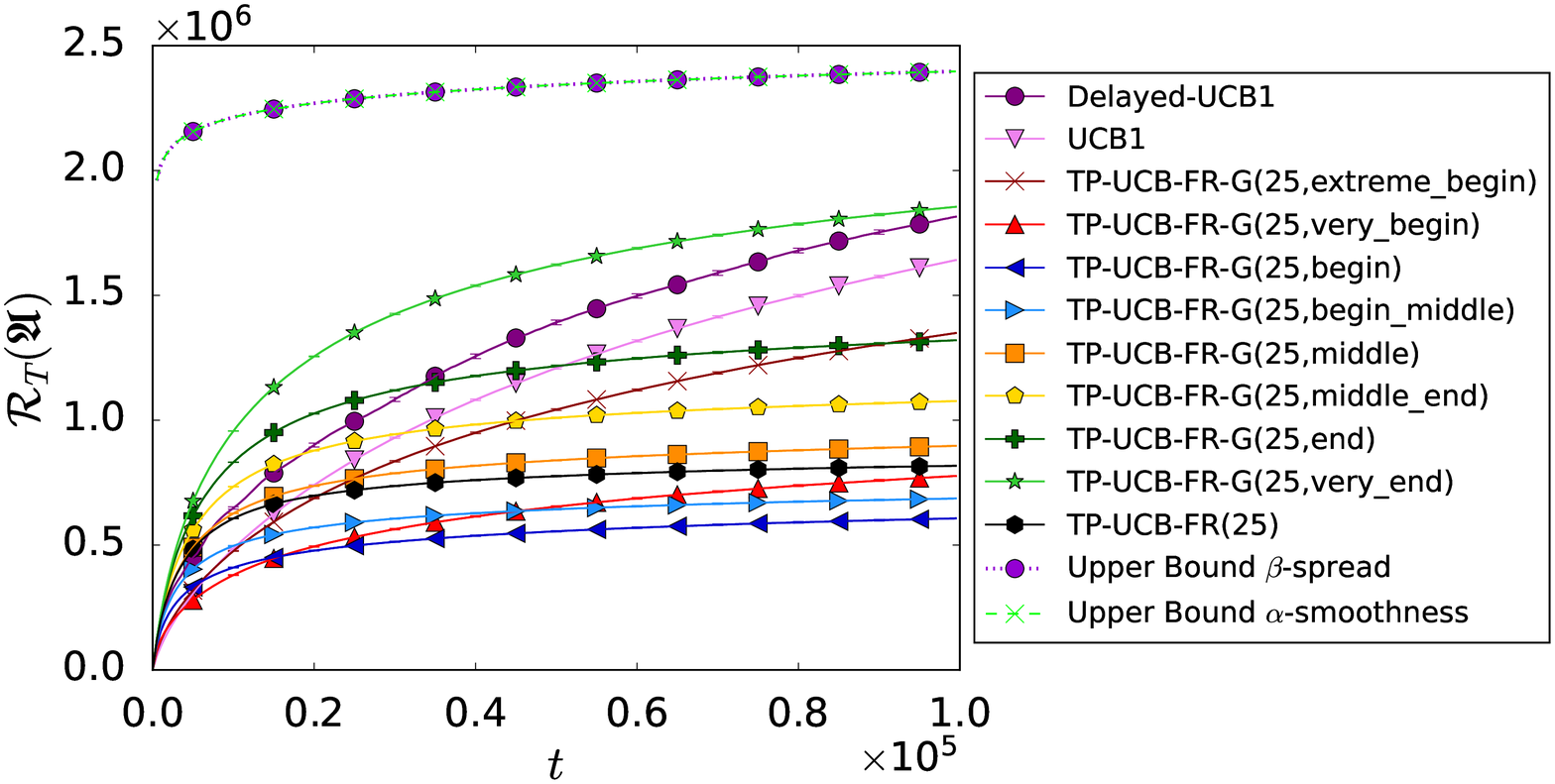}
  \caption{Regret against time for Setting 1 with $\alpha_{est} = 25$}
  \label{fig:setting1alpha25}
\end{figure}

The primary goal of this setting is to discover the influence of adjusting $\alpha$ over \alg \ learners. We run this setting for different choices of $\alpha$, $\alpha_{est} \in [5, 10, 20, 25, 50]$. In this setting, we model $k = 10$ arms, the reward is collected over $\tau_{\max} = 100$ rounds, and the maximum reward is set to be $\overline{R}^i = 100i$. The aggregate rewards for the learner proposed in \cite{Romano2022} are s.t. $Z^{i}_{t,k} \sim \frac{\overline{R}^i}{\alpha} U[0,1]$. \alg \ learners in this setting \textit{assume} a different aggregate rewards structure such that $Z^{i}_{t,k} \sim \frac{\overline{R}^i}{\alpha}Beta(a^i_k, b^i_k)$. The experiment is run over a time horizon $T = 10^5$, and the $\alpha$-smoothness constant is $\alpha = 20$. The results are averaged over $100$ independent runs.\\
Section \ref{sec:ExpResults} provides the results for $\alpha_{est} = 50$. 
In this section, let us focus on the results of \alg \ when $\alpha_{est} = [5, 10, 20, 25]$. These results are plotted in Figures \ref{fig:setting1alpha5}, \ref{fig:setting1alpha10}, \ref{fig:setting1alpha20} and \ref{fig:setting1alpha25} respectively. These results align with the conclusions from Section \ref{sec:ExpResults}. That is, as $\alpha_{est}$ increases, begin-centered learners perform increasingly better than other learners. For $\alpha_{est} = 5$, we see that the \texttt{begin\_middle} learner performs better for larger $t$. This indicates that, in case of lower $\alpha_{est}$ values, \texttt{begin} is too optimistic.

\begin{table}
    \centering
    \begin{tabular}{lllll}
        \toprule
        $\alpha$ & Regret & Regret & $\delta^+$\\
        & \texttt{TP-UCB-FR} & \alg & (\%) \\
        \midrule
        5 & $1.34 \times 10^6$ & $1.28 \times 10^6$ & 4.6*\\
        10 & $1.04 \times 10^6$ & $9.27 \times 10^5$ & 10.9 \\
        20 & $8.56 \times 10^5$ & $6.71 \times 10^5$ & 21.6 \\
        25 & $8.18 \times 10^5$ & $6.07 \times 10^5$ & 25.8 \\
        50 & $7.21 \times 10^5$  & $4.64 \times 10^5$ & 35.6 \\
        \bottomrule
    \end{tabular}
    \caption{Summary of performance gains by \alg \ in Setting 1. *For $\alpha_{est} = 5$ we show the \texttt{begin\_middle} distribution as this is the only configuration where it performs better than the \texttt{begin} distribution implicitly assumed elsewhere.}
    \label{tab:setting1performance}
\end{table}

In Table \ref{tab:setting1performance} we present a complete summary of the performance gains $\delta^+$ for Setting 1 using the \texttt{begin} distribution\footnote{The data for the other distributions within Setting 1 is provided in the supplementary material}. Note that $\delta^+$ denotes the decrease of average regret in percentages with respect to the regret of \texttt{TP-UCB-FR}.

\subsubsection*{Setting 2}
In the second setting, the effect of different (non-uniform) data generating distributions for the delayed partial rewards is evaluated. Furthermore, configurations with a higher $\tau_{\max}$ are tested. We model $K = 10$ arms again with aggregate rewards after an arm pull s.t.  $Z^{i}_{t,k} \sim \frac{\overline{R}^i}{\alpha}Beta(a^i_k, b^i_k)$. The time horizon is set to $T = 10^5$. We evaluate four configurations and three scenarios. The four configurations are related to $\tau_{\max}$ and $\alpha$. The three scenarios differ from each other because of the distribution of partial rewards after an arm pull. The first scenario has a uniform aggregate reward distribution and is equal to Setting 1. The remaining two have higher rewards at the end of the $\tau_{\max}$ interval (Setting 2.1) and higher rewards just after the arm pull (Setting 2.2), respectively. The vectors $a^i$ and $b^i$ represent all values $a^i_k$ and $b^i_k$ for $k \in \{1, \dots, \alpha\}$ and can be found in the referenced tables \ref{tab:config1}, \ref{tab:config2}, \ref{tab:config3} and \ref{tab:config4}. 

\begin{itemize}
    \item \textbf{Configuration 1}: $\tau_{\max} = 100, \alpha = 10$, see Table~\ref{tab:config1}
    \item \textbf{Configuration 2} $\tau_{\max} = 100, \alpha = 50$, see Table~\ref{tab:config2}
    \item \textbf{Configuration 3} $\tau_{\max} = 200, \alpha = 20$, see Table~\ref{tab:config3}
    \item \textbf{Configuration 4} $\tau_{\max} = 200, \alpha = 100$, see Table~\ref{tab:config4}
\end{itemize}






\begin{table}
    \centering
    \begin{tabular}{ll}
        \toprule
        Setting & Parameter vector \\
        \midrule
        & $a^i$\\
        Uniform & $\boldsymbol{1}_\alpha$ \\
        Setting 2.1 
        & [2,4,6,8,10,10,10,10,10,10]
        \\     
        Setting 2.2 
        & [10,10,10,10,10,10,8,6,4,2] 
        \\
        &\\
        & $b^i$ \\
        Uniform & $\boldsymbol{1}_\alpha$ \\
        Setting 2.1 & [10,10,10,10,10,10,8,6,4,2] \\
        Setting 2.2 & [2,4,6,8,10,10,10,10,10,10] \\
        \bottomrule
    \end{tabular}
    \caption{Distribution parameters for different scenarios with Configuration 1}
    \label{tab:config1}
\end{table}

\begin{table}
    \centering
    \begin{tabular}{ll}
        \toprule
        Setting & Parameter vector \\
        \midrule
        & $a^i$\\
        Uniform & $\boldsymbol{1}_\alpha$ \\
        Setting 2.1 
        & [2,4, \ldots, 48,50, \ldots, 50]
        \\     
        Setting 2.2 
        & [50, \ldots, 50,48, \ldots, 4,2] 
        \\
        &\\
        & $b^i$ \\
        Uniform & $\boldsymbol{1}_\alpha$ \\
        Setting 2.1 & [50, \ldots, 50,48, \ldots, 4,2] \\
        Setting 2.2 & [2,4, \ldots, 48,50, \ldots, 50] \\
        \bottomrule
    \end{tabular}
    \caption{Distribution parameters for different scenarios with Configuration 2}
    \label{tab:config2}
\end{table}





\begin{table}
    \centering
    \begin{tabular}{ll}
        \toprule
        Setting & Parameter vector \\
        \midrule
        &  $a^i$\\
        Uniform & $\boldsymbol{1}_\alpha$ \\
        Setting 2.1 
        & [2,4, \ldots, 18,20, \ldots, 20]
        \\     
        Setting 2.2 
        & [20, \ldots, 20,18, \ldots, 4,2]
        \\
        &\\
        &  $b^i$ \\
        Uniform & $\boldsymbol{1}_\alpha$ \\
        Setting 2.1 & [20, \ldots, 20,18, \ldots, 4,2] \\
        Setting 2.2 & [2,4, \ldots, 18,20, \ldots, 20] \\
        \bottomrule
    \end{tabular}
    \caption{Distribution parameters for different scenarios with Configuration 3}
    \label{tab:config3}
\end{table}




\begin{table}
    \centering
    \begin{tabular}{ll}
        \toprule
        Setting & Parameter vector \\
        \midrule
        & $a^i$\\
        Uniform & $\boldsymbol{1}_\alpha$ \\
        Setting 2.1 
        & [2,4, \ldots, 98,100, \ldots, 100]
        \\     
        Setting 2.2 
        & [100, \ldots, 100,98, \ldots, 4,2]
        \\
        &\\
        & $b^i$ \\
        Uniform & $\boldsymbol{1}_\alpha$ \\
        Setting 2.1 & [100, \ldots, 100,98, \ldots, 4,2] \\
        Setting 2.2 & [2,4, \ldots, 98,100, \ldots, 100] \\
        \bottomrule
    \end{tabular}
    \caption{Distribution parameters for different scenarios with Configuration 4}
    \label{tab:config4}
\end{table}

\subsubsection*{Results}
\begin{figure}
  \includegraphics[width=\linewidth]{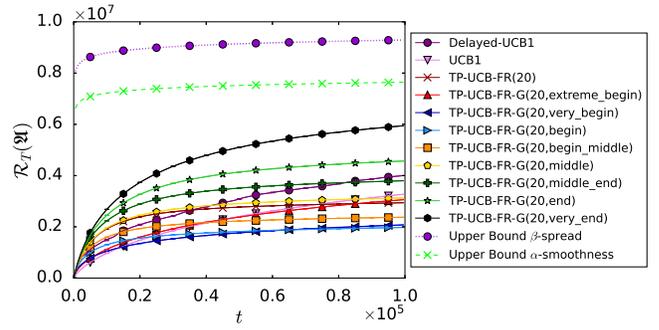}
  \caption{Regret against time for Setting 2.1 with $\tau_{\max} = 200, \alpha_{est} = 20$}
  \label{fig:setting2.1.200.20}
\end{figure}

\begin{figure}
  \includegraphics[width=\linewidth]{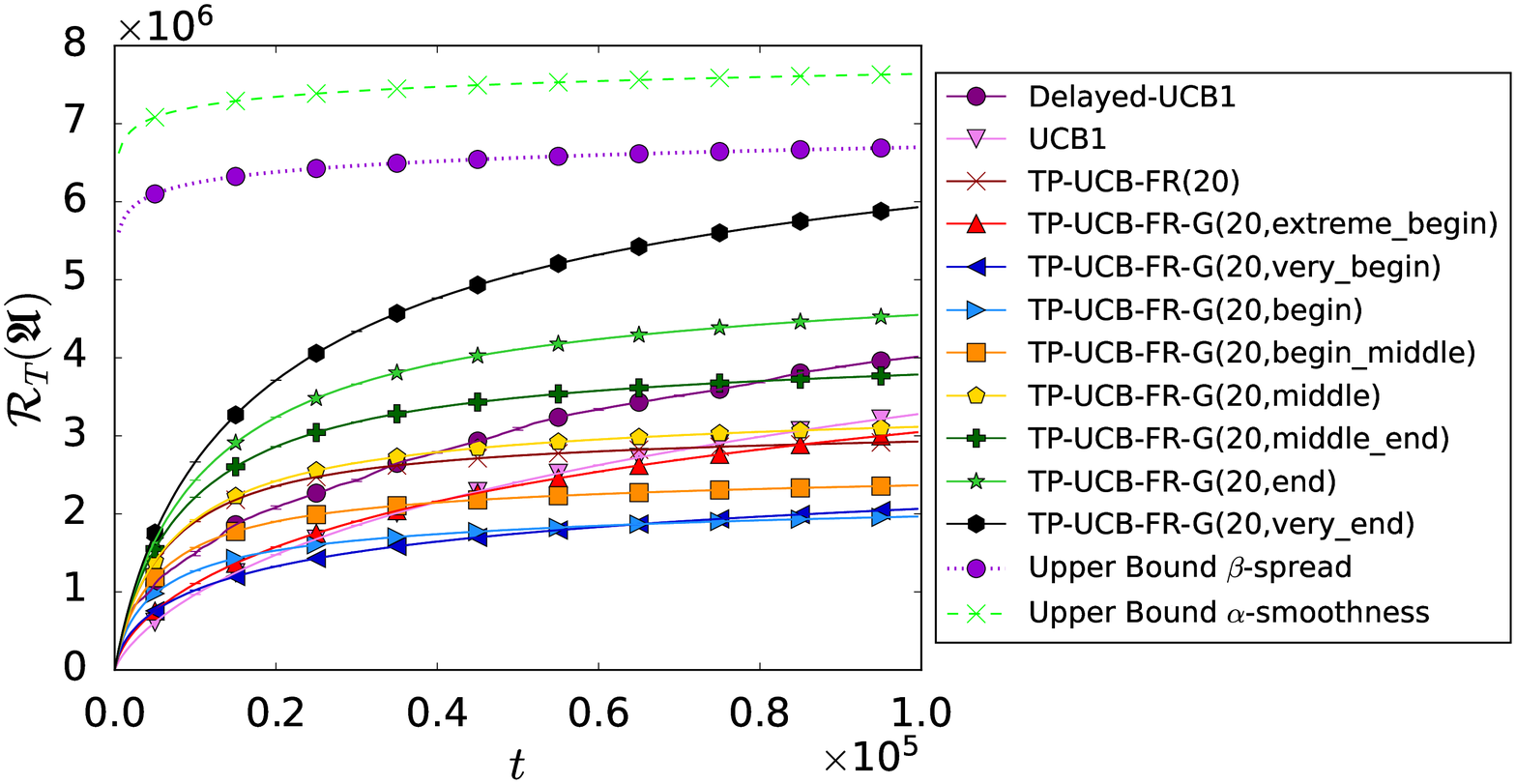}
  \caption{Regret against time for Setting 2.2 with $\tau_{\max} = 200, \alpha_{est} = 20$}
  \label{fig:setting2.2.200.20}
\end{figure}

\begin{figure}
  \includegraphics[width=\linewidth]{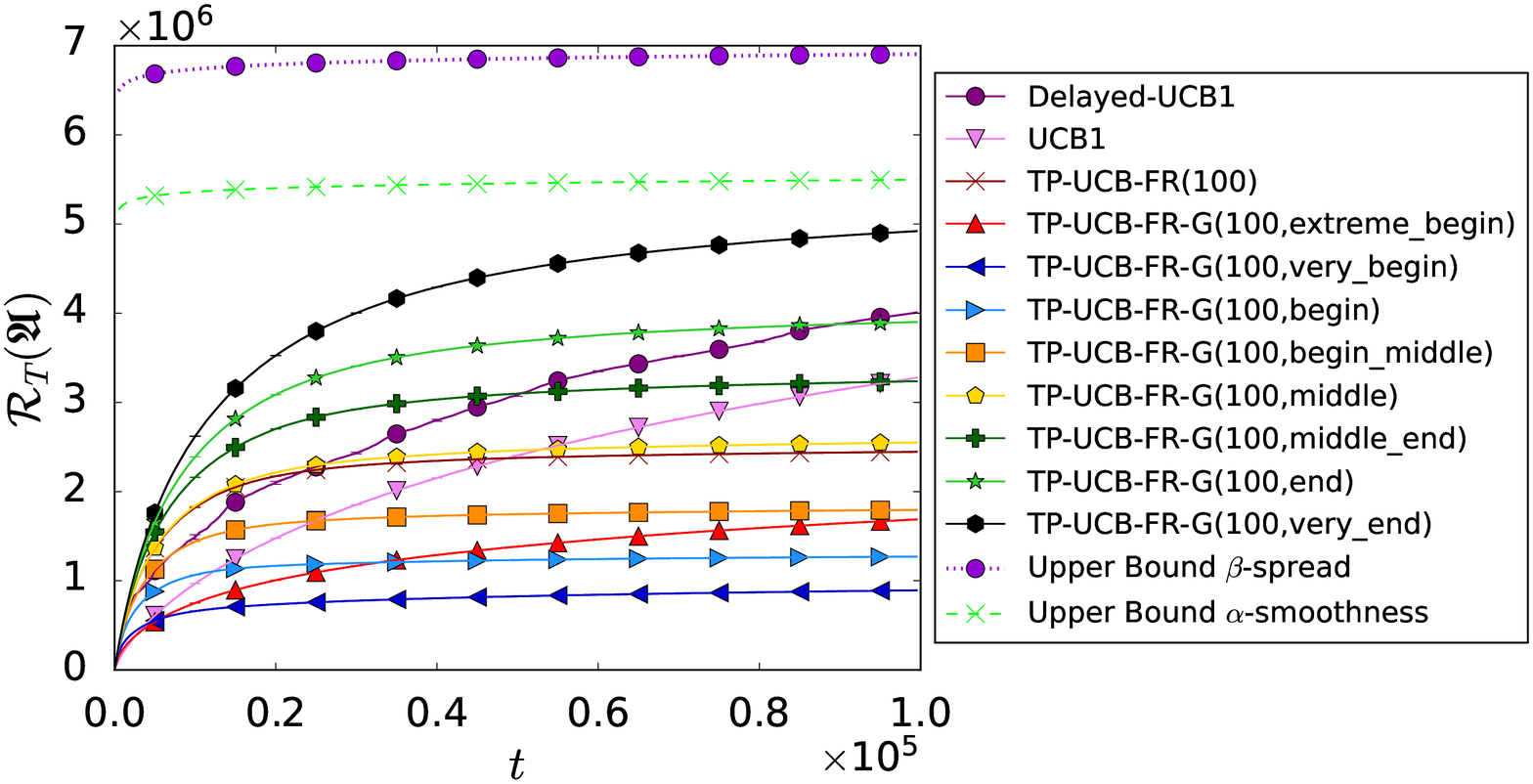}
  \caption{Regret against time for Setting 2.1 with $\tau_{\max} = 200, \alpha_{est} = 100$}
  \label{fig:setting2.1.200.100}
\end{figure}

\begin{figure}
  \includegraphics[width=\linewidth]{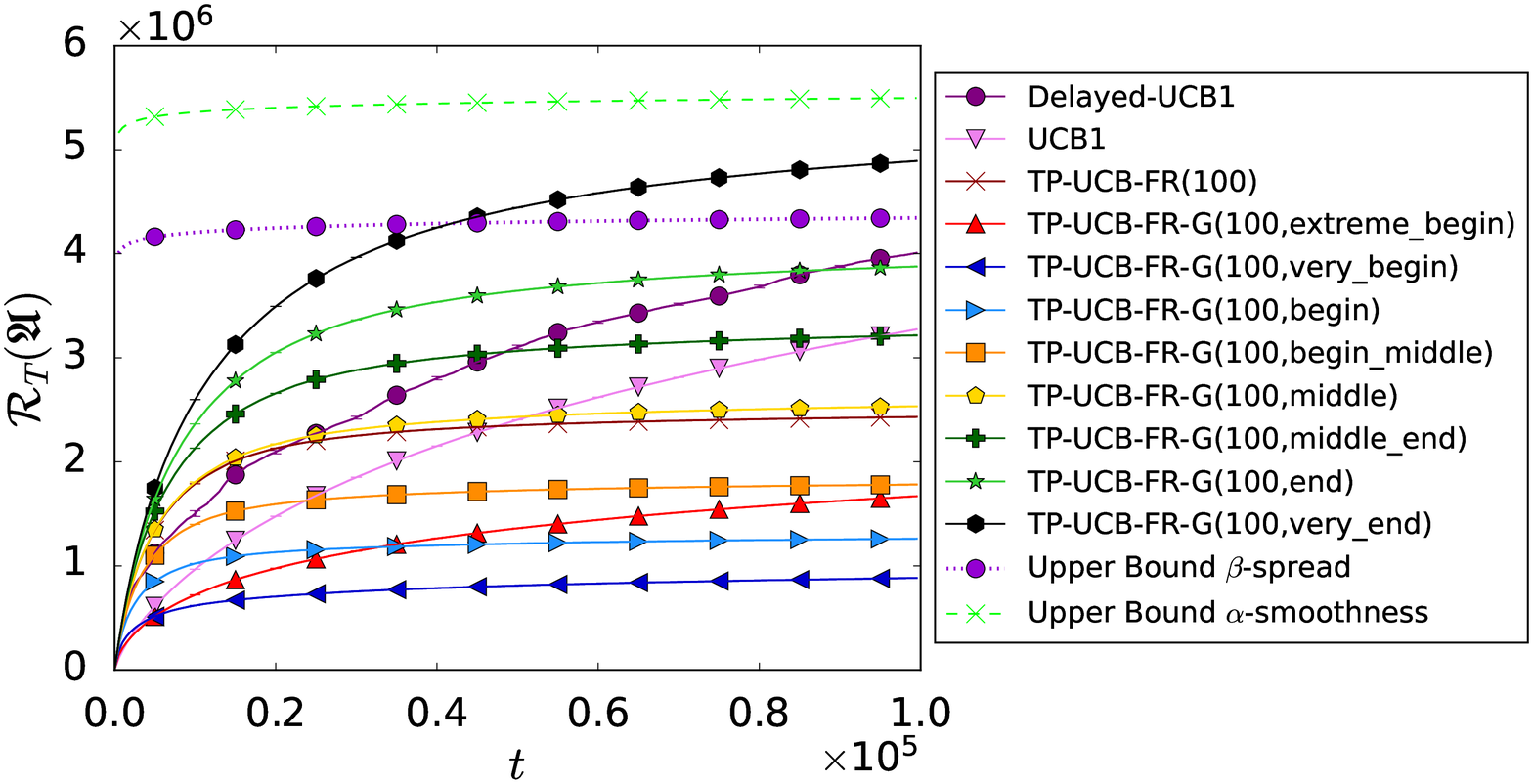}
  \caption{Regret against time for Setting 2.2 with $\tau_{\max} = 200, \alpha_{est} = 100$}
  \label{fig:setting2.2.200.100}
\end{figure}

As discussed in the main paper in Section~\ref{sec:ExpResults}, the results for different scenarios within one configuration are visually indistinguishable. The analysis of Figure \ref{fig:diffs_1_3} in the same section shows that there are no significant differences between learner increases. It also shows that all learners perform best in Setting 2.2, worse in the uniform setting and worst in Setting 2.1. Because no visual distinction can be made, the results for Configuration 1 and Configuration 2 are visually identical to the results for Setting 1 with $\alpha_{est}=10$ and $\alpha_{est}=50$. Similarly, the results of the Setting 2 tests with uniform distribution are visually identical to the results of Setting 2.1 and 2.2. To show the change of our upper bound compared to the upper bound of \texttt{TP-UCB-FR}, we present the results of running Configuration 3 and 4 for Setting 2.1 and 2.2. The resulting plots can be seen in Figures \ref{fig:setting2.1.200.20}, \ref{fig:setting2.2.200.20}, \ref{fig:setting2.1.200.100} and \ref{fig:setting2.2.200.100}. Furthermore, in Tables \ref{tab:setting2.1performance}, \ref{tab:setting2.2performance} and \ref{tab:setting2uniformperformance}, we provide a complete summary of the performance gains $\delta^+$ for different configurations in Setting 2 using the \texttt{begin} distribution\footnote{The data for the other distributions within Setting 2 is provided in the supplementary material}.

\begin{table}
    \centering
    \begin{tabular}{lllll}
        \toprule
        $\tau_{\max}$ & $\alpha$ & Regret & Regret & $\delta^+$\\
        & & \texttt{TP-UCB-FR} & \alg & (\%) \\
        \midrule
        100 & 10 & $1.04 \times 10^6$ & $9.27 \times 10^5$ & 10.9 \\
        100 & 50 & $7.19 \times 10^5$ & $4.66 \times 10^5$ & 35.2 \\
        200 & 20 & $2.94 \times 10^6$ & $1.98 \times 10^6$ & 32.7 \\
        200 & 100 & $2.45 \times 10^6$  & $1.27 \times 10^6$ & 48.2 \\
        \bottomrule
    \end{tabular}
    \caption{Summary of performance gains $\delta^+$ by \alg \ in Setting 2.1}
    \label{tab:setting2.1performance}
\end{table}

\begin{table}
    \centering
    \begin{tabular}{lllll}
        \toprule
        $\tau_{\max}$ & $\alpha$ & Regret & Regret & $\delta^+$\\
        & & \texttt{TP-UCB-FR} & \alg & (\%) \\
        \midrule
        100 & 10 & $1.04 \times 10^6$ & $9.25 \times 10^5$  & 11.0 \\
        100 & 50 & $7.17 \times 10^5$ & $4.64 \times 10^5$  & 35.3  \\
        200 & 20 & $2.93 \times 10^6$ & $1.97 \times 10^6$ &  32.8 \\
        200 & 100 & $2.43 \times 10^6$ & $1.26 \times 10^6$ & 48.1 \\
        \bottomrule
    \end{tabular}
    \caption{Summary of performance gains $\delta^+$ by \alg \ in Setting 2.2}
    \label{tab:setting2.2performance}
\end{table}

\begin{table}
    \centering
    \begin{tabular}{lllll}
        \toprule
        $\tau_{\max}$ & $\alpha$ & Regret & Regret & $\delta^+$\\
        & & \texttt{TP-UCB-FR} & \alg & (\%) \\
        \midrule
        100 & 10 & $1.04 \times 10^6$ & $9.28 \times 10^5$ & 10.8 \\
        100 & 50 & $7.18 \times 10^5$ & $4.65 \times 10^5$ & 35.2 \\
        200 & 20 & $2.94 \times 10^6$ & $1.98 \times 10^6$ & 32.7 \\
        200 & 100 &$ 2.44 \times 10^6$ & $1.27 \times 10^6$ & 48.0\\
        \bottomrule
    \end{tabular}
    \caption{Summary of performance gains $\delta^+$ by \alg \ in Setting 2 uniform}
    \label{tab:setting2uniformperformance}
\end{table}

\subsubsection{Spotify Setting}

The Spotify setting is run with the same pre-processing and configuration detailed in \cite{Romano2022} to accurately compare the performance between their \texttt{TP-UCB-FR} algorithm and our proposed \alg \ algorithm. We average the results over 100 independent runs. The results of this setting can be seen in Figure \ref{fig:settingspotify}.

\subsection{Additional Experiments}\label{App:additionalexperiments}

\subsubsection{Alternative distributions as input for \alg}
\begin{figure}
  \includegraphics[width=\linewidth]{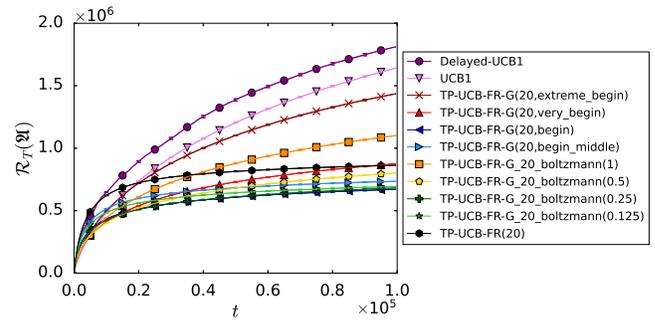}
  \caption{Boltzmann and Beta-Binomial distribution comparison for Setting 1 with $\alpha_{est} = 20$}
  \label{fig:boltzmann}
\end{figure}

In Section \ref{sec:ExpResults}, we describe the results obtained by various settings, using Beta-Binomial distributions with different choices for their parameters. In general, the domain of a distribution must be bounded on the interval $[1, \alpha]$ in order to be used as input. Below, we describe several finite discrete distributions that can be restricted to have such a domain, and the outcomes of the experiments.\\\\
\textit{The Zipfian distribution}\\
\begin{figure}
  \includegraphics[width=\linewidth]{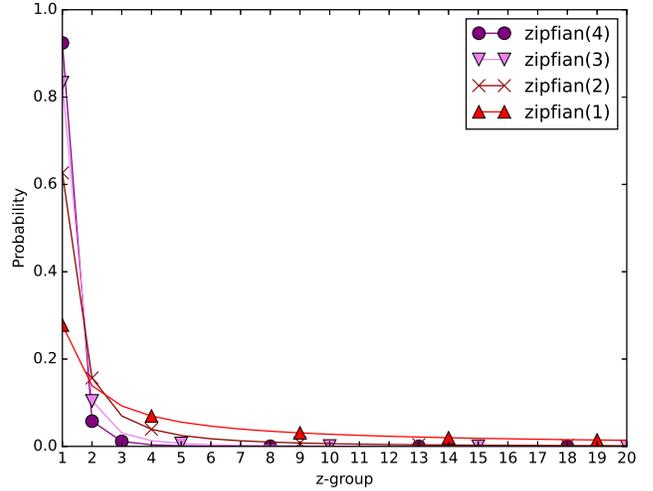}
  \caption{PMF of Zipfian distributions with different parameters}
  \label{fig:zipfianpmf}
\end{figure}
The Zipfian distribution allows us to describe 'begin-oriented' distributions with different parameters, as seen in Figure \ref{fig:zipfianpmf}. Since we observe that begin-oriented distributions pair well with our algorithm, Zipfian distributions should perform well in this setting.\\
Let us consider Setting 1. We run an experiment for $\alpha_{est} = 20$, which is a perfect estimation of the smoothness factor. Furthermore, we only consider learners that are begin-oriented. We add 4 learners using Zipfian distributions with parameters $s \in [1, 2, 3, 4]$ with names \texttt{zipfian(1)}, \texttt{zipfian(2)}, \texttt{zipfian(3)} and \texttt{zipfian(4)} respectively.
\begin{figure}
  \includegraphics[width=\linewidth]{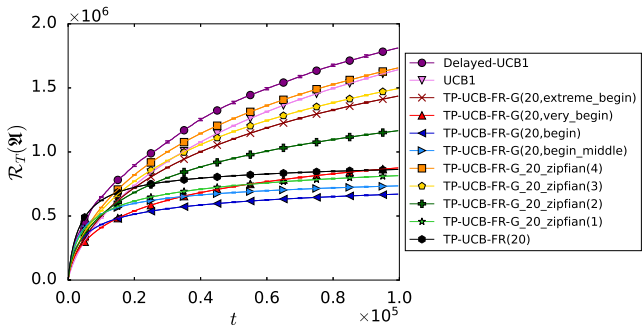}
  \caption{Zipfian and Beta-Binomial distribution comparison for Setting 1 with $\alpha_{est} = 20$}
  \label{fig:zipfian}
\end{figure}
We observe that the learner \alg\_zipfian(1) performs better than the learner proposed by \cite{Romano2022}, but worse than a begin-oriented Beta-Binomial learner. These results show that zipfian$(\zeta)$ learners for $\zeta \geq 2$ are outperformed by most Beta-Binomial learners, and do not bring any improvements to the results.\\\\
\textit{The Boltzmann distribution}\\
\begin{figure}
  \includegraphics[width=\linewidth]{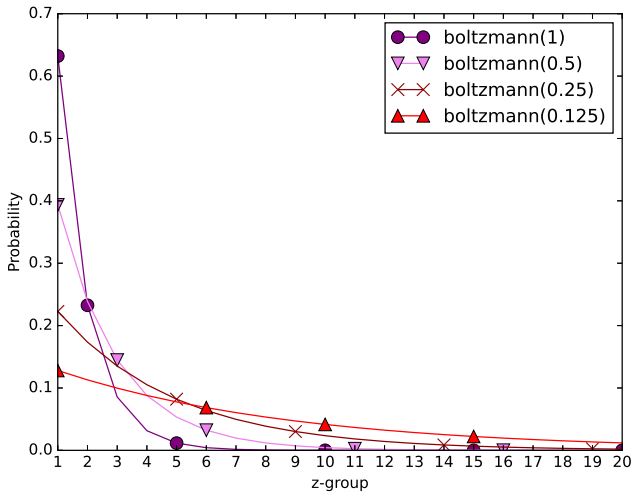}
  \caption{PMF of Boltzmann distributions with different parameters}
  \label{fig:boltzmannpmf}
\end{figure}
In a similar way, we can also describe 'begin-oriented' distributions using the Boltzmann distribution. We denote the input distribution as \texttt{boltzmann}$(\lambda)$. We notice that as $\lambda \to 0$, the distribution provides a smoother begin-orientation. This experiment is particularly interesting because for $\lambda = 0.25$ and $\lambda = 0.125$, the distribution is carefully begin-centered, so it could result in a performance gain.
We consider again Setting 1. We run an experiment for $\alpha_{est} = 20$, representing a perfect estimation of the smoothness factor. We consider the Beta-Binomial learners that are begin-oriented, and add Boltzmann distribution learners for $\lambda \in [1, 0.5, 0.25, 0.125]$. The results for this experiment are shown in Figure \ref{fig:boltzmann}.
We observe that Boltzmann distributions, in particular \texttt{Boltzmann(0.5)}, provide marginally better performance for $t = [0, 15000]$, and is nearly identical to the curve of the \texttt{very\_begin} learner. This is only logical, since the curves in Figures \ref{fig:parameters} and \ref{fig:boltzmannpmf} are nearly identical. It is also important to observe that for $t = [0, 38000]$, \texttt{Boltzmann(0.25)} provides marginally better performance than \texttt{begin}. For $t > 38000$, \texttt{begin} is still the learner with the lowest regret. Since $\lambda = 0.125$ starts to under-estimate, and $\lambda = 1, \lambda = 0.5$ are clear over-estimations, we have sufficient evidence that this distribution does not provide the required flexibility needed for performance gains over Beta-Binomial distributions. Precise regret values for this experiment, can be found in the supplementary material. We exclude \texttt{extreme\_begin} and \texttt{Boltzmann(1)} because they are clear over-estimations, and \texttt{Delayed-UCB}, \texttt{UCB} because they are irrelevant for this comparison.

\begin{figure}
  \includegraphics[width=\linewidth]{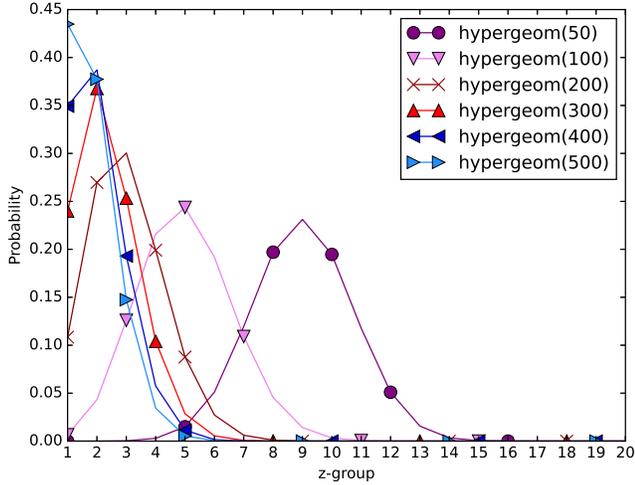}
  \caption{PMF of Hypergeometric distributions with different parameters}
  \label{fig:hypergeompmf}
\end{figure}

With the above results we conclude that Beta-Binomial distributions still give the best performance and flexibility, whilst Boltzmann distributions lack in both areas mentioned.\\\\
\textit{The Hypergeometric distribution}\\
For a mathematical description of the Hypergeometric distribution, we refer to \cite{lee_2012}. In short, a random variable $X$ follows a hypergeometric distribution if its probability mass function is defined as 
\begin{equation}
    p_X(k) = P(X=k) = \frac{\binom{K}{k}\binom{N-K}{n-k}}{\binom{N}{n}}
\end{equation}
where $N$ is the population size, $K$ the number of success rates in the population, $n$ the number of draws and $k$ the number of successes for which the PMF returns the probability. We set $n$ and $K$ to $\alpha-1$, equal to the number of $z$-groups minus one and shift each value of $k$ given as input by 1 (similarly to our Beta-Binomial implementation), such that we bound $p_X(k)$ to the domain $[1, \alpha]$. We can then shape the distribution with some choice for $N \geq 2\alpha$. In this section, we denote learners using the Hypergeometric distribution as \texttt{Hypergeom}(N) where the shaping parameter $N \in \mathbb{N}$ and $N \geq 2\alpha = 40$. We shall describe several begin-oriented hypergeometric distributions for $N \in [50, 100, 200, 300, 400, 500]$. Figure \ref{fig:hypergeompmf} depicts the corresponding probability mass functions for $N$.

\begin{figure}
  \includegraphics[width=\linewidth]{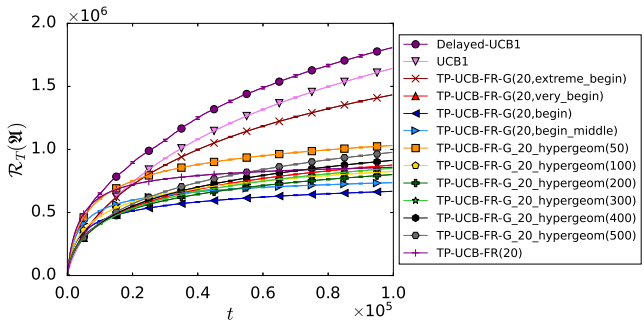}
  \caption{Hypergeometric and Beta-Binomial distribution comparison for Setting 1 with $\alpha_{est} = 20$}
  \label{fig:hypergeom}
\end{figure}

We shall include begin-oriented beta-binomial learners for comparison, and run Setting 1 with $\alpha_{est} = 20$, a perfect smoothness factor estimation. The results for this experiment are shown in Figure \ref{fig:hypergeom}

Again, we observe that \alg$(20, \texttt{begin})$ is the best performing learner when averaged over the entire time horizon $T$. However, for $t \in [0, 15000]$, \texttt{hypergeom}$(N)$ with $N \in [200, 300, 400, 500]$, learners with such distributions perform marginally better. Due to the slope of the corresponding curves created by hypergeometric distributions, their performance degrades as $t$ gets larger. Due to the large gaps between $N$, one could search for the most appropriate value for $N$ and perhaps discover a distribution that performs better than the \texttt{begin} Beta-Binomial learner, but this requires running Setting 1 for $200 \leq N \leq 500$ which is very costly. Moreover, we could fine-tune the Beta-Binomial learner in a similar way, by running Setting 1 for different $\alpha, \beta$ in the proximity of \texttt{begin}, which will result in far less runs. Again, the Beta-Binomial distribution remains superior in flexibility and performance.

\subsubsection{Relation between assumed distribution and upper bound}

As mentioned in Theorem \ref{Thm:UpperBound}, the upper bound on \alg \ learners only holds when the assumed distribution that is given as input to the algorithm matches the data generating distribution. In this section, we want to go into more detail about this and discuss which type of learner, with a non-matching assumed distribution, stays below this upper bound.

We note that this phenomenon, although not mentioned explicitly in \cite{Romano2022}, is also present for the upper bound of the \texttt{TP-UCB-FR} algorithm and that it depends on the $\alpha$ of the data generating distribution and not on the estimate given as input. Because the assumed distribution for this algorithm is univariate (only depending on $\alpha$), determining which learners do not exceed the upper bound is tested more easily: an increased $\alpha_{est}$ always leads to better results in all tested settings (see section \ref{sec:ExpResults}). This means that learners typically stay below the upper bound when $\alpha_{est} \in [\alpha, \tau_{\max}]$. 

However, the upper bound of \alg \ depends on three properties of the data generating distribution, namely its expected value, the index of coincidence and $\alpha$. In the results (see section \ref{sec:ExpResults}), we see that learners with a lower assumed expected value generally perform better. But there is a limit to how low the expected value can be, because the second factor, the index of coincidence, gets relatively high for very begin oriented distributions which raises the regret. This makes it harder to give an empirical estimate about which kind of learners stay below the theoretical upper bound. It also shows that learners with a higher $\alpha$ perform better, similarly to the case of \texttt{TP-UCB-FR}. Deriving the precise correlation between these three parameters is highly complex and it is a potential future work. 

\end{document}